\documentclass{article} 

\usepackage{microtype}
\usepackage{graphicx}
\usepackage{subcaption}
\usepackage{booktabs} 

\usepackage{hyperref}
\usepackage{url}

\usepackage[preprint]{icml2026}

\usepackage{amsmath}
\usepackage{amssymb}
\usepackage{mathtools}
\usepackage{amsthm}

\usepackage[capitalize,noabbrev]{cleveref}

\theoremstyle{plain}

\theoremstyle{definition}

\theoremstyle{remark}

\icmltitlerunning{}

\usepackage{multirow}
\usepackage{todonotes}
\usepackage{xfrac}
\usepackage{enumitem}

\usepackage{amsmath,amsfonts,bm}

\def\eqref#1{equation~\ref{#1}}

\def\1{\bm{1}}

\DeclareMathAlphabet{\mathsfit}{\encodingdefault}{\sfdefault}{m}{sl}
\SetMathAlphabet{\mathsfit}{bold}{\encodingdefault}{\sfdefault}{bx}{n}

\usepackage{caption}  
\usepackage{textcomp}
\usepackage{wrapfig}

\usepackage{algorithm}

\usepackage[table]{xcolor}
\newcommand{\methodname}{\textsc{MASH}}

\newcommand{\HYPERPARAMETER}{\item[\textbf{Hyperparameter:}]}
\newcommand{\RETURN}{\textbf{return} }

\begin{document}

\twocolumn[
  \icmltitle{MASH: Modeling Abstention via Selective Help-Seeking}

  \icmlsetsymbol{equal}{*}

  \begin{icmlauthorlist}
    \icmlauthor{Mustafa Omer Gul}{yyy}
    \icmlauthor{Claire Cardie}{yyy}
    \icmlauthor{Tanya Goyal}{yyy}
  \end{icmlauthorlist}

  \icmlaffiliation{yyy}{Department of Computer Science, Cornell University, Ithaca, NY. USA}

  \icmlcorrespondingauthor{Mustafa Omer Gul}{mog29@cornell.edu}

  \icmlkeywords{Machine Learning}

  \vskip 0.3in
]

\printAffiliationsAndNotice{} 

\begin{abstract}
\addtocounter{footnote}{1}
LLMs cannot reliably recognize their parametric knowledge boundaries and often hallucinate answers to outside-of-boundary questions. In this paper, we introduce \methodname~(\textbf{M}odeling \textbf{A}bstention via \textbf{S}elective \textbf{H}elp-seeking), a training framework that readily extracts abstentions from LLMs. Our key idea is that any external help-seeking by an LLM, i.e. search tool use, can serve as a proxy for abstention if the external help (search) is appropriately penalized while also rewarding answer accuracy. \methodname~operationalizes this idea using reinforcement learning with a pay-per-search reward. We run experiments on three knowledge-intensive QA datasets. Our results show that \methodname~ substantially improves upon the selective help-seeking performance of prior efficient search approaches; on multi-hop datasets, it improves answer accuracy by $7.6\%$. Furthermore, \methodname~demonstrates strong off-the-shelf abstention performance, showcasing behavior competitive with prior abstention methods that additionally require predetermining model knowledge boundaries to construct training data. Overall, we show \methodname~training effectively aligns search tool use with parametric knowledge, which can be successfully leveraged for making abstention decisions and efficient search tool use. \footnote{Code and checkpoints are available at \url{https://github.com/momergul/mash}.}

\end{abstract}

\section{Introduction}
\vspace{-2mm}
A reliable AI assistant should recognize its knowledge boundaries, what questions it can and cannot effectively respond to, and act accordingly when a question is outside its boundaries. Conventionally, LLMs learn their knowledge boundaries through alignment by explicitly training for abstention~\citep{yang2024alignment, cheng2024can} and calibrated verbalization of uncertainty~\citep{xu-etal-2024-sayself, stengel-eskin2024lacie}. These strategies yield improved recognition of capability boundaries but are limited to reducing model errors. The number of questions a model can correctly answer remains unchanged. In this paper, we ask if we can design a training strategy that yields an abstention model capable of recognizing its boundaries, while 
learning techniques that expand its set of answerable questions?

We look at human behavior for inspiration. Humans recognize their limitations and can abstain when asked for knowledge they lack. Alternatively, they can seek outside help to answer these otherwise unanswerable questions. In this paper, we propose \methodname~(\textbf{M}odeling \textbf{A}bstention via \textbf{S}elective \textbf{H}elp-seeking), a framework that indirectly trains LLMs for abstention by instead training models to engage in selective help-seeking, i.e. asking for help only when it cannot effectively respond to a query alone. 

\begin{figure*}[t]
    \centering
    \includegraphics[scale=0.24,trim=0mm 190mm 150mm 10mm]{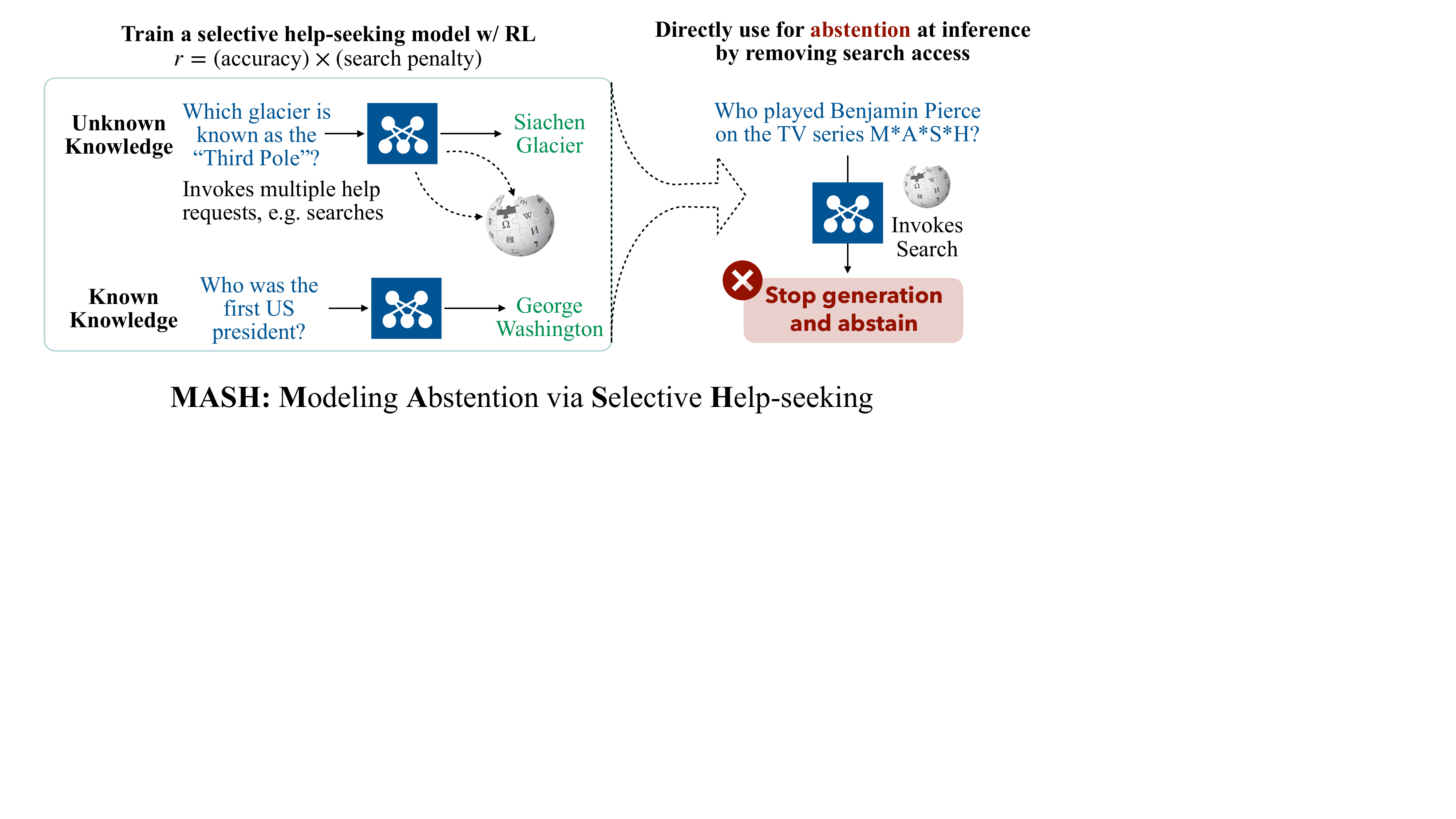}
    \vspace{-1mm}
    \caption{Overview of \methodname's strategy for abstention. Help-seeking LLMs are RL-trained to maximize accuracy while minimizing searches. At inference, this same model is used for abstention by removing search access and treating any search requests as abstention.}
\label{fig:mash} \vspace{-4mm}
\end{figure*}

As a proof of concept, we explore this idea in the context of short-form question-answering tasks. We operationalize help-seeking as invoking a retrieval tool that returns information related to a given query. We train LLMs that selectively seek help (i.e.~invoke retrieval) end-to-end with reinforcement learning using a penalty that discounts a correctness reward by the number of searches a model performs. An optimal policy optimizing this reward would, by definition, search only when a question cannot be reliably answered with parametric knowledge. In an inference mode with the same access to search, this model will mirror the above selective search behavior. More importantly, we can readily elicit abstention decisions from this same model by removing access to search tools. In that case, any search invocation serves as a proxy for abstention (see Figure~\ref{fig:mash}). 
\methodname, under this framing, effectively trains for two capabilities at the cost of one. 
Crucially, \methodname~assumes no privileged information regarding knowledge boundaries like standard abstention approaches~\citep{yang2024alignment, cheng2024can, xu-etal-2024-sayself} or require structured multi-agent interactions~\citep{stengel-eskin2024lacie, eisenstein2025dont}.

We train \methodname~models using reinforcement learning with a pay-per-search reward (see Figure~\ref{fig:mash}). However, baseline implementations of this idea \citep{wang2025acting} result in efficient but sub-optimal search behaviors: models can converge to always searching at least once. To address this, we propose a lightweight synthetic data curation and SFT pipeline that, crucially, assumes no information about knowledge boundaries. Instead, it serves to inject diverse, albeit parametrically unaligned, search behavior to improve exploration in RL training. Additionally, we extend reward formulations of prior work \citep{wang2025acting} to obtain penalties with harsher levels of severity; we found this crucial for extracting good help-seeking behaviors via RL.

We run our experiments on 3 different knowledge-intensive datasets, and evaluate both the selective help-seeking performance with regular inference (with access to search) and abstention performance (without access to search). Our results show that \methodname~models substantially outperform previous efficient search baselines~\citep{wang2025acting} at balancing answer accuracy and searches. Notably, on multi-hop datasets, \methodname~reports a $7.6\%$ accuracy improvement with a better distribution of searches. In fact, this performance is on par with search baselines~\citep{jin2025searchr} that allow any number of searches (up to a max value) without any penalty. We investigate this further and show that this improvement can be attributed to \methodname~showcasing a broader range of search strategies, i.e. diversity over number of searches, as a direct result of its training recipe.

Furthermore, we show that \methodname~reports strong off-the-shelf abstention performance. It achieves competitive performance with our strongest abstention baseline DPO \citep{rafailov2023direct, cheng2024can}, which explicitly constructs a specialized training dataset for abstention training. Moreover, compared to prompting and supervised training methods for abstention~\citep{yang2024alignment}, \methodname~reports higher answer accuracy ($10-20\%$ improvement) over non-abstained questions by better differentiating between answerable/unanswerable questions. 

Taken together, our results demonstrate that \methodname~is an effective technique that yields an abstention model capable of recognizing its boundaries, while simultaneously expanding its set of answerable questions via help-seeking.

\section{\methodname: Modeling Abstention via Selective Help-seeking}

\subsection{Abstention Framework}    
\label{sec:problem_defn}
\vspace{-2mm}
\paragraph{Help-seeking LLMs} We assume an inference setting where a language model $\pi_\theta$ can ask for help by sending a help request $h$ to a helper $H(\cdot)$, which then returns a response $o \sim H(h)$. This helper $H$ can take various forms: it could be a tool such as a retrieval model responding to a query, another stronger language model or an actual human in-the-loop. The model would then condition on the response $o$ and continue its generation. Formally, given an input question $q$, the model samples a trajectory $\tau \sim \pi_\theta(\cdot | q; H)$ of the form 
    $\tau = (r_1, h_1, o_1, \cdots, r_l, h_l, o_l, r_{l+1}, \hat{y}) $, where each $r_i$ represents reasoning, each $h_i$ represents a help request generated by $\pi_\theta$, $o_i$ represents the associated output from helper $H(\cdot)$ and $\hat{y}$ represents the model's final answer.

In this paper, we focus on knowledge-based domains. Here, $h_i$ is a search query generated by $\pi_\theta$, the helper $H(\cdot)$ is a retrieval model and $o_i$ is a set of top-$k$ documents retrieved by $H(h_i)$ from a document corpus. In practice, we assume that reasoning outputs $r_i$ are enclosed between $<$think$>$ and $<$/think$>$, search queries between $<$search$>$ and $<$/search$>$, and answers between $<$answer$>$ and $<$/answer$>$ tokens. We use the terms help/search, and helper/retriever interchangeably. 

\textbf{Training Objective} \hspace{2mm}We want the language model $\pi_\theta$ to recognize its knowledge boundaries. We posit that we can obtain such a model, without privileged information regarding parametric knowledge boundaries, by training the model to maximize its accuracy while minimizing the number of search requests. Specifically, we optimize the following proxy objective:
\begin{align}\label{eq:proxy_objective}
    \max_\theta ~ &\mathbb{E}_{(q, y) \sim D, \tau \sim \pi_\theta(\cdot | q; H)}[r_{acc}(y, \tau) \cdot r_{help}(q, \tau)] \\
    &- \beta D_{KL}[\pi_\theta(\tau | q; H) || \pi_{\theta_{init}}(\tau | q; H)], \nonumber
\end{align}
where $D$ is the dataset, $r_{acc}(y, \tau) \in \{0, 1\}$ is a binary measure of correctness and $r_{help}(q, \tau) \in [0, 1]$ is a multiplicative penalty assigning a lower value the greater the number of searches in $\tau$ is.  We use reinforcement learning, with the GRPO algorithm~\citep{guo2025deepseek}, for optimization. 

\textbf{Eliciting Abstention from a Selectively Help-Seeking Model}\hspace{2mm} Let $\pi_{\theta^*}$ be the optimal policy derived using the above objective. This model will selectively seek help as a function of its parametric accuracy for a given question $q$ and the severity of the $r_{help}$ penalty. Particularly, we prove in Appendix \ref{sec:theoretical_analysis} that $\pi_{\theta^*}$ will always answer a question $q$ parametrically, without help, if and only if its expected parametric accuracy for $q$ is greater than the maximum attainable expected reward for $q$ when seeking help, which relies on $r_{help}$. This is intuitive, as the optimal policy should choose to seek help for question $q$ only if this resulted in an equivalent or higher reward than answering parametrically.

We re-frame the goal (and our subsequent evaluations) of our help-seeking model from efficiency, i.e. reducing number of searches, to parametric knowledge alignment, i.e. aligning search behavior with presence or absence of knowledge about a given question in the model's parameters. Under this re-framing, we elicit abstentions from a selectively help-seeking model by treating any search invocation as a proxy for abstention. Following from our theoretical analysis in Appendix \ref{sec:theoretical_analysis}, the optimal policy $\pi_{\theta^*}$ would abstain on a question $q$ if and only if its expected parametric accuracy is less than or equal to the reward it would achieve if it were to seek help. Figure~\ref{fig:mash} illustrates this abstention framework, which we call \methodname: \textbf{M}odeling \textbf{A}bstentions via \textbf{S}elective \textbf{H}elp-seeking.

\subsection{Training a Selective Help-Seeking Model}
\methodname~training involves two main steps: (1) initializing $\theta_{init}$ in Equation~\ref{eq:proxy_objective} such that it displays diverse search behaviors (zero, one, or multiple searches) to encourage exploration, and (2) a reward function that appropriately balances accuracy and search tool penalty.

\subsubsection{Initializing $\pi_\theta$ w/ Warm-start SFT}
\label{sec:naive_sft}



RL training to optimize Equation \ref{eq:proxy_objective} should, in theory, result in a model that selectively seeks help. However, in practice, we find that such training converges to sub-optimal policies that either exhibit degenerate strategies that always or never search, or that fail to learn to use the search tool effectively. In our work, we propose a \textbf{lightweight and model-agnostic synthetic data generation and finetuning pipeline} that results in a substantially better initial policy for subsequent RL training. Our data generation pipeline is designed to encourage diversity in the number of searches in model trajectories. Crucially, it requires no information about models' parametric knowledge boundaries. 

\textbf{Synthetic data generation} \hspace{2mm} Our overall algorithm is a simple constrained decoding procedure. Let $G$ be the synthetic data generation model. For each input question $q$ in the warm-start training dataset, we first randomly sample a target number of searches $l \sim \{0, \ldots, l_{max}\}$ to perform, where $l_{max}$ is a hyperparameter controlling the maximum number of allowed searches. We then manually set the action tag (e.g., $<$think$>$, $<$search$>$ or $<$answer$>$) the model should output at the start of each turn given the target number of searches and let the model perform generation normally otherwise. We sample $N$ such trajectories per question, evaluate each and preferentially return a correct trajectory. Note that this constrained decoding process is only used during synthetic data generation. Refer to Algorithm ~\ref{alg:warm_start} in Appendix \ref{sec:app_warm_start} for more details.

A warm start SFT step is also included in recent works' training pipelines to improve subsequent RL training~\citep{guo2025deepseek, gandhi2025cognitive, wang2025octothinker}. However, we highlight one key difference. Contrary to prior works, our warm start process does not target correctness or alignment with models' parametric knowledge, the two central goals of \methodname. In fact, our synthetic data contains $35\%$ errors with respect to answer correctness and, by design through the random sampling of search counts, yields a policy whose search behavior is unaligned with its parametric knowledge (discussed  in Appendix~\ref{sec:app_warm_start}). We further mitigate any confounding factors by intentionally using a generator model $G$ different from the model to be trained (as discussed in Section \ref{sec:experiments}). The model learns how and when to use searches during RL training.

\subsubsection{Reward Formulation}
\label{sec:penalty_defn}
Our reward $r(y, \tau)$ is a product of two terms: $r_{acc}(y, \tau)$, which is a binary correctness reward and $r_{help}(q, \tau)$, which is a search tool penalty. We compute $r_{acc}(y, \tau)$ using exact match. The form and severity of $r_{help}$ will influence the learned help-seeking behavior. For input question $q$ and $G$ output trajectories $\{\tau_i\}^G_{i=1}$ sampled during GRPO, let $n$ be the number of search queries in the most efficient and correct trajectory $\tau^\text{ef}$ and $m$ be the number of queries in the given trajectory $\tau_i$. We want $r_{help}$ to appropriately penalize $\tau_i$ if $m > n$. There exists an arbitrarily high number of penalty formulations that satisfy this desiderata; we experiment with three: 

1.~\textbf{Exponential Decay}, defined as $r^{\mathrm{EXP}}_{help}(q, \tau_i) = \lambda^{m-n}$ where $\lambda$  controls the severity of the penalty.

2. \textbf{OTC} reward proposed by \citet{wang2025acting}. We follow their recommendation and set $c$ to the maximum number of searches allowed in a single trajectory. \vspace{-2mm}
\begin{equation}
    r^\text{OTC}_{help}(q, \tau_i) = \begin{cases}
        1 & \text{if } m = n = 0 \\
        \cos(\frac{m \cdot \pi}{2m + c}) & \text{if } n = 0 \\
        \sin(\frac{m \cdot \pi}{m + n}) & \text{otherwise} \vspace{-2mm}
    \end{cases},
\end{equation}

3. \textbf{OTC-Strict} which enforces an extremely strict penalty when $m > n=0$. Note that $n=0$ indicates there is a correct trajectory $\tau^\text{ef}$ without searches. For these cases, any other trajectory $\tau_i$ that uses searches should get no reward under a very strict definition of answerability. Therefore, we set $r^\text{OTC-St}_{help}(q, \tau_i)$ to 0 for such cases. We can use any of the above two reward formulations for when $n > 0$, but choose OTC's sinusoidal function to align with prior work.

\section{Experimental Setup}
\label{sec:experiments}\vspace{-2mm}
\textbf{Datasets and Models} \hspace{2mm} We run our experiments on three knowledge-intensive datasets: the single-hop dataset Natural Questions (NaturalQA)~\citep{kwiatkowski2019natural}, and multi-hop datasets HotPotQA~\citep{yang-etal-2018-hotpotqa} and 2WikiMultiHopQA (2Wiki)~\citep{ho-etal-2020-constructing}.\footnote{We find that the ``comparison'' and ``bridge-comparison'' questions comprising in 2WikiMultiHopQA have unbalanced answer distributions (skewed towards ``no''). This opens up the possibility of reward hacking by exploiting this dataset property. Therefore, we omit these questions from our training and evaluation.} We train and evaluate on each dataset separately; this allows us to evaluate \methodname~across tasks requiring different search strategies and with different distributions of parametrically answerable questions. Particularly, NaturalQA and HotPotQA differ in search strategies but have similar distributions of parametrically answerable questions, while 2Wiki is a multi-hop dataset containing a greater proportion of unanswerable questions than both (see Appendix \ref{sec:app_datasets} for further analysis). We perform all training and evaluation on the Qwen2.5-3B base model~\citep{qwen2025qwen25technicalreport}. We deliberately choose the base model over instruct as the latter has already undergone abstention training although the exact training strategy is unknown; we propose \methodname~as an alternative. We use the E5 retriever~\citep{wang2022text} and the 2018 Wikipedia dump as our knowledge source~\citep{karpukhin-etal-2020-dense}. 

\textbf{Hyperparameters}   
For the OTC reward, we follow \citet{wang2025acting} and set $c$ equal to the maximum number of searches. For Exponential Decay, we set $\lambda$ to $0.5$ for Natural Questions and $0.8$ otherwise, following hyperparameter tuning. We note that these hyperparameter choices imply the following decreasing order of severity of search penalty: \textsc{otc-strict}$\rightarrow$\textsc{exp}$\rightarrow$\textsc{otc}. For each search query, we fix the response to be the top-3 retrieved passages and allow a maximum of $5$ searches per trajectory. We use the veRL library~\citep{10.1145/3689031.3696075} for RL training.  More training details are in Appendix~\ref{sec:app_grpo_training}.

\textbf{Warm-start data generation} We follow the strategy outlined in Section \ref{sec:naive_sft} to generate warm-start data for each dataset using Qwen2.5-32B base. Choosing a separate model ensures that information about knowledge boundaries is not baked into the SFT training data, while using the base variant avoids abstention behavior present in the instruct model from entering the training data and confounding results. For each dataset, we randomly sample $1000$ questions from its training set and set $l_{max}=2$. We select the trajectory for each question from $N=5$ samples. Details can be found in Appendix~\ref{sec:app_warm_start}

We evaluate our selective help-seeking models in two inference modes: (1) \textbf{with access to search tools}, which directly aligns with its training, and (2) \textbf{without search tools}, where we use the help-seeking model for abstention. The baselines and evaluation metrics for these are described next.

\subsection{Details for Inference Mode I: w/ Search Tools}\label{sec:setup_tools} \vspace{-2mm}
\paragraph{Baselines} We compare \methodname's help-seeking model against the following baselines that also conduct RL training, but with different setups:  (1) \texttt{\textbf{R1}} trained using RL but without access to any search tools during training or evaluation. This baseline provides an upper bound for answer accuracy using only parametric knowledge. (2) \texttt{\textbf{Search-R1}}~\citep{jin2025searchr} trained with search tools and a binary correctness reward; showcasing an upper bound without any penalties for searching. We set the maximum number of searches for Search-R1 to 3 due to compute and memory concerns. (3) \texttt{\textbf{OTC}}~\citep{wang2025acting} RL-trained for efficient search tool use. 
We compare these baselines to three \methodname~variants that differ in reward penalties (refer to \S~\ref{sec:penalty_defn}). Note that \methodname~w/ OTC and OTC differ in the warm-start procedure applied to the former.

\textbf{Evaluation Metrics} \hspace{2mm}We want our help-seeking model to strike a balance between answering parametrically (without search calls) and seeking help (with search calls). We report three metrics that collectively capture this: (1)~\textbf{Accuracy (Acc)}, i.e. if the predicted answer matches the gold response. Due to the limitations of exact match, we use an LLM judge, namely DeepSeek-V3.1~\citep{liu2024deepseek}, to determine this. (2)~\textbf{Tool calls (TC)}, i.e. the average number of searches across trajectories. (3)~\textbf{Tool Productivity (TP)} \citep{wang2025acting}, which is defined as $[\sum_{i=1}^{|\mathcal{D}|} \sfrac{\mathbb{I}\{y_i = \hat{y}_i\}}{(1 + m_i)}]/|\mathcal{D}|$ for test set $\mathcal{D}$. This discounts the accuracy of each output trajectory by its number of searches $m_i$. For all models, we report these metric averages over $4$ samples. 
We use TP on the validation set to select our model checkpoints for all methods, except Search-R1 for which we use accuracy; TP results in a much inferior checkpoint selection for this case.

\begin{table*}[t]
\centering
\small
\caption{Accuracy, average number of tool calls (TC) and tool productivity (TP) for baselines and \methodname~evaluated under \textbf{inference w/  search tools}. \methodname~w/ OTC-ST is our best model with a $4.22\%$ and $5.62\%$ mean improvement on Acc and TP over baseline OTC.}\vspace{-2mm}
\begin{tabular}{l|ccc|ccc|ccc}
\toprule
\multirow{2}{*}{\textbf{Method}} & \multicolumn{3}{c|}{\textbf{Natural Questions}} & \multicolumn{3}{c|}{\textbf{HotPotQA}} & \multicolumn{3}{c}{\textbf{2Wiki}} \\ 
\cmidrule(lr){2-4} \cmidrule(lr){5-7} \cmidrule(lr){8-10} 
& Acc$\uparrow$ & TC$\downarrow$ & TP$\uparrow$ & Acc$\uparrow$ & TC$\downarrow$ & TP$\uparrow$ & Acc$\uparrow$ & TC$\downarrow$ & TP$\uparrow$  \\
\midrule
R1 & 26.04 & 0.0 & 26.04 & 26.53 & 0.0 & 26.53 & 9.18 & 0.0 & 9.18 \\
Search-R1 \citep{jin2025searchr} & 57.31 & 1.0 & 28.66 & 56.34 & 3.00 & 14.09 & 45.39 & 2.98 & 11.35 \\
OTC \citep{wang2025acting} & 58.91 & 1.0 & 29.45 & 44.76 & 0.81 & 28.64 & 39.59 & 1.57 & 15.32 \\ \midrule
\methodname~w/ \sc{otc}  & 59.81 & 1.0 & 29.95 & 55.42 & 1.14 & \textbf{32.91} & 45.99 & 1.6 & 18.87 \\
\methodname~w/ \sc{otc-st}  & 56.37 & 0.64 & \textbf{38.63} & 53.32 & 1.10 & 32.55   & 46.23 & 1.64 & \textbf{19.08} \\
\methodname~w/ \sc{exp}  & 54.35 & 0.65 & 36.61  & 53.78 & 1.07 & 32.09 & 44.32 & 1.53 & 18.10 \\
\bottomrule
\end{tabular}%
\vspace{-5mm}
\label{tab:iid_tool_use}
\end{table*}

\subsection{Details for Inference Mode II: Abstention}\label{sec:setup_abstention}\vspace{-1mm}
In this mode, we follow the \methodname~process outlined in Figure~\ref{fig:mash} and \S~\ref{sec:problem_defn} to extract abstentions from a help-seeking model by removing access to search tools at inference. 

\textbf{Baselines} \hspace{2mm}We compare against the following baselines: \vspace{-2mm} 
\begin{enumerate}[leftmargin=*, noitemsep,topsep=0pt] 
\item \textbf{5-shot prompting} with the base model, with abstention/not of in-context exemplars decided based on its parametric knowledge. 
\item \textbf{Alignment for Honesty - Absolute} (AFH-Abs)~\citep{yang2024alignment}, which does SFT on a specially curated abstention dataset by pairing each input question with either the output ``I abstain'' or the gold answer, depending on the base model's knowledge boundaries. 
\item \textbf{Alignment for Honesty - Multisample} (AFH-Mult)~\citep{yang2024alignment}, which constructs multiple training samples for each question, pairing it with either ``I abstain'' or the gold answer depending on the average correctness over multiple outputs, for SFT training.
\item \textbf{DPO}, inspired by \citet{cheng2024can}, which pairs each question with a preferred and dispreferred output. If the question is parametrically answerable, we set these to be the gold answer and ``I abstain'' respectively; this is switched for unanswerable questions. We train with the DPO loss objective~\citep{rafailov2023direct} and SFT loss added as a regularizer \citep{pang2024iterative}. 
\item \textbf{Ternary}, as considered by~\citet{xu2024rejection} and~\citet{wei2025truthrl}, which conducts RL training with GRPO using a ternary reward assigning $+1$ for correct answers, $0$ for abstention, and $-1$ abstention. We use the exact match reward from our other RL baselines to assess correctness.
\end{enumerate}

{Each of (1), (2) and (4) requires a definition of answerability; i.e.~when we can claim that a question is answerable. A standard technique is to estimate the accuracy over $10$ samples and use a threshold $\eta$ to classify into answerable or not. However, there does not exist a consensus in prior works on how to decide this threshold~\citep{yang2024alignment, cheng2024can}. In our paper, we follow \cite{yang2024alignment} and set $\eta=0.1$.  
Exact data curation and training details are in Appendix~\ref{sec:app_abstention_details}.}

\textbf{Evaluation Metrics} For abstention evaluation, we report two kinds of metrics: (1) \textbf{Answer Accuracy}: We report overall Accuracy, i.e.~over the test set, and Precision, i.e.~over non-abstained questions. Note that over-conservativeness, i.e. aggressively abstaining, will hurt overall accuracy but help precision, while under-conservativeness will cause the opposite. (2) \textbf{Abstention Classification}: This captures whether a model's abstention behavior is aligned with its knowledge boundaries, agnostic of accuracy. To avoid defining answerability (different reward penalties assume a different threshold), we evaluate over two groups of questions unaffected by the choice of $\eta$: questions that the base models always answer incorrectly, with an average accuracy of $0$, or always correctly, with an average accuracy of $1$. Let $\text{\%Abs}(0)$ and $\text{\%Abs}(1)$ be the percentage of questions for which a model abstains for the above two groups, respectively. We report $\%\text{Abs}(0)$ and Delta ($\%\text{Abs}(0) - \%\text{Abs}(1)$). A model recognizing its knowledge boundaries should have a high abstention rate for always incorrect questions, i.e.  $\text{\%Abs}(0)$, and a much lower rate for always correct questions, captured by a large margin $\%\text{Abs}(0) - \%\text{Abs}(1)$. We do not evaluate on 2Wiki for abstention classification as there are only $58$ test examples in the $\text{Abs}(1)$ bucket, preventing reliable conclusions.

\section{Results}
\label{sec:results}
\vspace{-2mm}

\subsection{Inference Mode I: w/ Search Tools}
\label{sec:search_tool_results}
\vspace{-2mm}
We first evaluate the performance of baselines and \methodname~ in the inference setting with access to search tools. Table \ref{tab:iid_tool_use} reports overall answer accuracy, average tool calls and tool productivity for all methods. Additionally, we show the distribution of tool calls (TC=0/1/2+) and the corresponding accuracy per search count (subscript) in Table~\ref{tab:iid_search_distribution}. This allows us to conduct an apples-to-apples comparison between models' accuracy for the same number of tool calls.

\begin{table*}[t]
\centering
\small
\caption{Fine-grained tool use distribution (TC=0/1/2+ search) for baseline OTC and \methodname. We also report answer accuracies for questions in each subset (subscript). For instance, \methodname~w/ OTC-Strict on 2Wiki answers questions with 2+ searches $77.5\%$ of time, with an accuracy of $49.2\%$. \methodname~successfully off-loads questions to parametric answering (from TC=1 to TC=0) with minimal or no decrease in accuracy (HotPotQA \& NaturalQA). Highlight intensity correlates with the proportion of questions answered with TC=0/1/2+.} \vspace{-2mm}
\setlength{\tabcolsep}{2.5pt}
\begin{tabular}{l|ccc|ccc|ccc}
\toprule
\multirow{2}{*}{\textbf{Method}} & \multicolumn{3}{c|}{\textbf{Natural Questions}} & \multicolumn{3}{c|}{\textbf{HotPotQA}} & \multicolumn{3}{c}{\textbf{2Wiki}} \\
\cmidrule(lr){2-4} \cmidrule(lr){5-7} \cmidrule(lr){8-10} 
& 0 & 1 & 2+ & 0 & 1 & 2+ & 0 & 1 & 2+  \\
\midrule
OTC  & $0.0_{0.0}$ & \cellcolor{blue!30}$100.0_{58.9}$ & $0.0_{0.0}$  & \cellcolor{blue!7}$19.5_{64.5}$ & \cellcolor{blue!25}$80.2_{40.0}$ & \cellcolor{blue!2}$0.3_{32.0}$  & \cellcolor{blue!5}$3.1_{24.1}$ & \cellcolor{blue!13}$36.7_{26.6}$ & \cellcolor{blue!22}$60.2_{48.3}$  \\
\methodname~w/ \sc{otc}  & \cellcolor{blue!2}$0.2_{53.6}$ & \cellcolor{blue!30}$99.8_{59.8}$ & $0.0_{33.3}$ & \cellcolor{blue!10}$23.5_{66.5}$ & \cellcolor{blue!15}$41.7_{58.2}$ & \cellcolor{blue!13}$34.8_{44.6}$ & \cellcolor{blue!10}$13.0_{31.3}$ & \cellcolor{blue!10}$13.9_{35.9}$ & \cellcolor{blue!25}$73.1_{50.5}$   \\
\methodname~~w/ \sc{otc-st} & \cellcolor{blue!13}$36.4_{57.4}$ & \cellcolor{blue!20}$63.5_{55.8}$ & \cellcolor{blue!2}$0.1_{17.6}$  & \cellcolor{blue!11}$28.9_{59.9}$ & \cellcolor{blue!13}$34.7_{56.4}$ & \cellcolor{blue!13}$36.4_{45.2}$    & \cellcolor{blue!10}$14.3_{32.5}$ & \cellcolor{blue!8}$8.3_{42.3}$ & \cellcolor{blue!25}$77.5_{49.2}$  \\
\methodname~~w/ \sc{exp}& \cellcolor{blue!10}$35.2_{53.6}$ & \cellcolor{blue!20}$64.8_{54.8}$ & $0.0_{20.0}$   & \cellcolor{blue!10}$23.7_{64.0}$ & \cellcolor{blue!18}$45.5_{53.4}$ & \cellcolor{blue!12}$30.8_{46.5}$   & \cellcolor{blue!8}$11.8_{32.2}$ & \cellcolor{blue!12}$23.4_{20.6}$ & \cellcolor{blue!20}$64.9_{55.1}$   \\
\bottomrule
\end{tabular}%
 \vspace{-4mm}
\label{tab:iid_search_distribution}
\end{table*}

\textbf{\methodname~outperforms all search baselines on tool productivity by effectively balancing accuracy and searches.}
Our results in Table~\ref{tab:iid_tool_use} show that \methodname, particularly \methodname~w/ OTC-Strict, leads to a $5.62$ point improvement on tool productivity over baseline OTC on average across datasets. Surprisingly, \methodname~variants report accuracies on par with Search-R1 (trained without any tool use penalty) on multi-hop datasets HotPotQA and 2Wiki, but with a substantially lower number of searches (e.g. $1.10/1.64$ vs $3/2.98$). Moreover, this performance is a massive improvement over baseline OTC ($8.56\%$ and $6.64\%$ improvements on HotPotQA and 2Wiki respectively) with only a slightly higher number of searches. Tool productivity, which accounts for both these metrics, improves by 3.83 points on average over baseline OTC. Together, these results suggest that \methodname~not only reduces the number of searches, but also better operationalizes them to maintain accuracy.

\textbf{Severe search penalties are needed for parametric answers for single-hop NaturalQA.} We observed that both baseline OTC and \methodname~with the lenient OTC penalty (\methodname~w/ OTC) do not learn to answer parametrically for NaturalQA, i.e. converge to TC=1 for all questions. On the other hand, \methodname~w/ OTC-Strict answers parametrically for $36\%$ of the questions with only a $2.5\%$ drop in accuracy, thereby improving tool productivity by $9$ points.  Similarly, \methodname~w/ Exp-Dec answers parametrically $35\%$ of the time, with a $4.5\%$ drop in accuracy\footnote{Note that \methodname~w/ Exp-Dec training did result in checkpoints with higher accuracies. However, we use tool productivity on the validation set as the metric to select the final checkpoint.} compared to baseline OTC but a 7 point improvement in tool productivity.

The multi-hop datasets, HotPotQA and 2Wiki, report slightly higher average tool calls with the strictest penalty (\methodname~w/ OTC-Strict), presumably contradicting the above claim. However, fine-grained search distributions (see Table~\ref{tab:iid_search_distribution}) show that,  similarly to NaturalQA, OTC-Strict does answer parametrically (TC=0) more often than the lenient versions. The increase in average tools calls is due to a larger fraction of 2 searches.

\textbf{\methodname~variants extract better and more diverse search behaviors for multi-hop datasets.} Comparing search statistics for \methodname~w/ OTC and baseline OTC in Table~\ref{tab:iid_search_distribution}, we see that they report a comparable number of parametric answers ($23.5\%$ vs $19.5\%$) but show very different search behaviors for the remaining questions.
Particularly, the baseline OTC model without warm-start collapses to only one search for the remaining $80.2\%$ of its trajectories, while the warm-started model (\methodname~w/ OTC) can perform a mixture of one and multi-hop searches. In fact, \methodname~variants report a much higher accuracy for one search questions ($56.4$\% vs $40.0$\%) by offloading the more ``difficult'' questions, i.e. those the model cannot answer with only one search, to the two search bucket. Baseline OTC fails to do this and reports lower overall accuracy. We see similar trends for the other multi-hop dataset, 2Wiki, as well.

\textbf{\methodname~successfully aligns search tool use with parametric knowledge.} For NaturalQA, the fine-grained search statistics in Table~\ref{tab:iid_search_distribution} show that the questions \methodname~w/ OTC-Strict and w/ Exp answer parametrically have similar answer accuracy compared to those for which they invoke one search call ($57.4$ vs $55.8$ for w/ OTC-Strict). This clearly shows that \methodname~can distinguish between parametrically answerable and not answerable questions and preferentially invoke tool calling for the latter to maintain overall accuracy.

\subsection{Inference Mode II: w/ Abstention}
\vspace{-2mm}

\textbf{\methodname~shows strong abstention behavior off-the-shelf.}
Table \ref{tab:iid_abstention} (left) reports the answer accuracy for the overall test dataset (Acc) and the non-abstained questions (Prec) for each method.\footnote{Note that it is possible to game one of these metrics by being over- or under-conservative. Therefore, all our conclusions are based on analyzing the two metrics collectively. } First, we observe that, apart from \methodname~w/ OTC on NaturalQA, all \methodname~variants substantially outperform the prompting and Alignment for Honesty based SFT approaches in terms of answer precision and report comparable overall accuracy. In a couple of instances, we find that the AFH (Absolute) baseline reports better accuracy (e.g. HotPotQA and NaturalQA) compared to \methodname, but this is accompanied by a 10-20\% drop in precision. 

We find that \methodname~w/ OTC-Strict, our best performing model from Section~\ref{sec:search_tool_results}, is comparable to DPO for NaturalQA and HotPotQA; it outperforms DPO based on Prec. ($59.9$ vs $53.1$ for HotPotQA) but reports lower overall accuracy ($17.3$ vs $19.9$). We attribute this to \methodname~w/ OTC-Strict being more conservative (i.e. more likely to abstain) than DPO. For 2Wiki, \methodname~w/ OTC-Strict outperforms DPO on both Acc and Prec. This competitive performance is despite \methodname~using the noisier exact match as the reward signal (due to our compute constraints), while DPO uses a stronger and more reliable LLM judge (DeepSeek-V3.1).

\begin{table*}[t]
\centering
\small
\caption{Answer accuracy and abstention classification results for specialized abstention approaches and \methodname. For answer accuracy, we report Acc over the entire test set and Prec, the accuracy over non-abstained answers. 
For classification, we report Abs(0), the abstention rate for unanswerable questions (higher better), and Delta, the margin between abstention rates for unanswerable and answerable questions (higher better). We highlight runs with similar performance but different trade-offs between Acc and Prec or Abs(0) and Delta in gray.} \vspace{-2mm}
\setlength{\tabcolsep}{4.5pt}
\begin{tabular}{l|cc|cc|cc||cc|cc}
\toprule
\multirow{2}{*}{\textbf{Method}} &  \multicolumn{6}{c||}{\textbf{Answer Accuracy}} & \multicolumn{4}{c}{\textbf{Abstention Classification}} \\
\cmidrule(lr){2-7} \cmidrule(lr){8-11}
& \multicolumn{2}{c|}{\textbf{NaturalQA}} & \multicolumn{2}{c|}{\textbf{HotPotQA}} & \multicolumn{2}{c||}{\textbf{2Wiki}} & \multicolumn{2}{c|}{\textbf{NaturalQA}} & \multicolumn{2}{c}{\textbf{HotPotQA}} \\
\cmidrule(lr){2-3} \cmidrule(lr){4-5} \cmidrule(lr){6-7} \cmidrule(lr){8-9} \cmidrule(lr){10-11}
& Acc & Prec  & Acc & Prec  & Acc & Prec  & $\text{Abs}(0)\uparrow$ & Delta$\uparrow$ & $\text{Abs}(0)\uparrow$ & Delta$\uparrow$   \\
\midrule
OTC  & 0.0 & -- & 12.6 & 64.5 & 0.75 & 24.1  & 100.0 & $0.0$ & 95.3 & 41.4 \\
\methodname~w/ \sc{otc}  & 0.1 & 53.6 & \cellcolor{gray!20}15.6 & \cellcolor{gray!20}66.5 & \cellcolor{gray!20}4.1 & \cellcolor{gray!20}31.3 & 99.9 & 0.1 & \cellcolor{gray!20}94.8 & \cellcolor{gray!20}52.3  \\
\methodname~w/ \sc{otc-st}  & \cellcolor{gray!20}20.9 & \cellcolor{gray!20}57.4 & \cellcolor{gray!20}17.3 & \cellcolor{gray!20}59.9 &\cellcolor{gray!20} 4.6 & \cellcolor{gray!20}32.5  & \cellcolor{gray!20}85.4 & \cellcolor{gray!20}66.1  & \cellcolor{gray!20}91.2 & \cellcolor{gray!20}60.3   \\
\methodname~w/ \sc{exp}  & 18.9 & 53.6 & \cellcolor{gray!20}15.2 & \cellcolor{gray!20}64.0 & \cellcolor{gray!20}3.8 & \cellcolor{gray!20}32.2 & 85.6 & 62.7  & \cellcolor{gray!20}94.5 & \cellcolor{gray!20}52.7  \\
\midrule
5-shot Prompting  & 23.4 & 42.5 & 14.7 & 31.5 & 3.6 & 10.9  & 60.2 & 44.6 & 60.5 & 26.9  \\
AFH (Absolute)  & 21.7 & 43.3 & 20.7 & 34.2 & 4.7 & 18.5 & 67.7 & 48.1 & 50.4 & 35.4 \\
AFH (Multisample)  & 14.7 & 54.8 & 12.9 & 53.8 & 2.6 & 29.2  & 87.9 & 52.1  & 89.2 & 57.6 \\
DPO  & \cellcolor{gray!20}22.3 & \cellcolor{gray!20}56.2 & \cellcolor{gray!20}19.9 & \cellcolor{gray!20}53.1 & \cellcolor{gray!20}3.3 & \cellcolor{gray!20}31.6  & \cellcolor{gray!20}84.5 & \cellcolor{gray!20}71.6  & \cellcolor{gray!20}85.9 & \cellcolor{gray!20}73.5  \\
Ternary & 0.0 & -- & 0.0 & -- & 0.0 & -- & 100.0 & 0.0 & 100.0 & 0.0 \\
\bottomrule
\end{tabular}%
\label{tab:iid_abstention} \vspace{-4mm}
\end{table*}

Finally, the Ternary baseline is outperformed by all methods and converges to always abstaining within 25 training steps on all datasets. This reproduces observations by~\citet{wei2025truthrl}, which similarly found convergence to abstaining for a vast majority of questions with exact match as the correctness signal. Using a stronger LLM judge alternative is outside the scope of this paper due to resource limitations.

\textbf{\methodname~can differentiate between answerable and unanswerable questions.}  Table~\ref{tab:iid_abstention} (right) shows the abstention classification results. As expected, we find that DPO models explicitly trained for abstention report the best results. Encouragingly, we see that \methodname~variants, except \methodname~w/ OTC on NaturalQA which does not learn to answer parametrically, report similarly high Abs(0) percentages as DPO. While DPO reports higher Delta for both datasets, Table~\ref{tab:iid_abstention} shows that these large improvements in Delta are often accompanied by a drop in precision. For example, DPO reports $13.2$\% better Delta than \methodname~w/ OTC-Strict for HotPotQA, but reports a $6.8\%$ lower precision.

Taken together, these results present an encouraging picture for the idea of modeling abstention with models trained for the auxiliary selective help-seeking task. They show that although \methodname~does not train explicitly for abstention, its \textbf{abstention behavior is analogous to that of abstention methods leveraging oracle information regarding model knowledge boundaries.}


\subsection{Generalization to Other Model Scales and Families}
\label{sec:model-generalization}\vspace{-2mm}
We repeat experiments on Qwen2.5-7B base and Qwen3-4B base to determine whether \methodname's success generalizes to different parameter scales and model families, respectively. Due to the cost of replicating all RL experiments, we restrict our attention to HotPotQA and conduct RL training only for the OTC baseline and for \methodname~w/ OTC-Strict, the best performing variant for Qwen2.5-3B base.

\textbf{\methodname~continues to outperform baselines at different model scales and families.} Under the inference setting with search enabled, the OTC baseline converges to searching once for both models. \methodname~w/ OTC-Strict, on the other hand, consistently performs a mixture of parametric answers, and one- and multi-hop searches. Furthermore, it achieves average improvements of $2.98\%$ and $9.59$ on accuracy and tool productivity. Similar trends persist for abstention. \methodname~w/ OTC-Strict outperforms prompting and Alignment for Honesty. \methodname~yet again displays competitive performance with DPO, with different accuracy-precision trade-offs. Particularly, Qwen2.5-7B-Base achieves superior precision ($60.67\%$ vs $48.25\%$), while Qwen3-4B-Base achieves superior accuracy ($20.96\%$ vs $16.72\%$) over DPO. Result tables can be found in Appendix \ref{sec:app_different_models}.

\subsection{Impact of Warm-start on Performance}
\label{sec:warm-start} \vspace{-2mm}

The comparative results of the OTC baseline and \methodname~w/ OTC in both Tables~\ref{tab:iid_tool_use} and \ref{tab:iid_search_distribution} indicate that the warm-start SFT training is key to \methodname's success. By design, it enables the model to explore diverse trajectories with varying numbers of search tool calls during RL. Here, we study the impact of warm start for all reward formulations. Table~\ref{tab:iid_tool_use_warm_start_ablation} reports the performance for all three without warm start (refer to Table~\ref{tab:iid_tool_use} for comparison with models trained with warm start). \vspace{-2mm}

\begin{table}[h]
\centering
\scriptsize
\caption{\methodname~\textbf{w/o warm-start} tested with search enabled. }\vspace{-2mm}
\setlength{\tabcolsep}{4pt}
\begin{tabular}{l|ccc|ccc|ccc}
\toprule
\multirow{2}{*}{\textbf{Method}} & \multicolumn{3}{c|}{\textbf{Natural Questions}} & \multicolumn{3}{c|}{\textbf{HotPotQA}} & \multicolumn{3}{c}{\textbf{2Wiki}} \\ 
\cmidrule(lr){2-4} \cmidrule(lr){5-7} \cmidrule(lr){8-10} 
& Acc$\uparrow$ & TC$\downarrow$ & TP$\uparrow$ & Acc$\uparrow$ & TC$\downarrow$ & TP$\uparrow$ & Acc$\uparrow$ & TC$\downarrow$ & TP$\uparrow$  \\
\midrule
\textsc{otc} & 58.91 & 1.0 & 29.45 & 44.76 & 0.81 & 28.64 & 39.59 & 1.57 & 15.32 \\ 
\sc{otc-st}  & 52.35 & 0.49 & 39.28 & \cellcolor{red!20}26.97 & \cellcolor{red!20}0.0 & \cellcolor{red!20}26.97 & \cellcolor{red!20}10.41 & \cellcolor{red!20}0.0 & \cellcolor{red!20}10.41 \\
\sc{exp}  & 57.59 & 1.00 & 28.80 & 41.46 & 0.71 & 28.67 & \cellcolor{red!20}9.72 & \cellcolor{red!20}0.0 & \cellcolor{red!20}9.72 \\
\bottomrule
\end{tabular}%
\label{tab:iid_tool_use_warm_start_ablation} \vspace{-5mm}
\end{table}

\textbf{Warm-start adds stability to harsher penalties.} The OTC reward shows the best help-seeking behavior when considering all datasets collectively. However, we discussed in \S~\ref{sec:search_tool_results} that the search behavior with warm-start is far superior to without for OTC. Recall that Exponential Decay and OTC-Strict both impose harsher penalties on search tool use than OTC. We observe that this results in severe training instabilities for these two when trained without warm-start -- the HotPotQA policy collapses to zero searches for OTC-Strict and the 2Wiki policy collapses for both Exponential Decay and OTC-Strict. Warm-start SFT, however, enables both to have successful training runs on all datasets, with OTC-Strict with warm-start even substantially outperforming OTC on all datasets for abstention metrics.

\subsection{Do oracle helpers improve selective help-seeking?}\label{sec:oracle_helper_main}\vspace{-2mm}
All experiments in Section~\ref{sec:results} rely on a retrieval model (E5; \cite{wang2022text}) as the helper $H(\cdot)$. However, search results output by these retrievers can be noisy, which in turn generates a noisy signal for training the selective help-seeking LLM via RL. 
This prompts us to investigate if improving the ``helper'', as opposed to the reward or initialization, can improve the learned help-seeking behavior. 

\textbf{Setup:} We set $H(\cdot)$ to be an oracle; it directly returns the gold answer if the LLM invokes a help tag in its trajectory (exact prompts used is included in Appendix \ref{sec:oracle_helper_appendix}). We train all \methodname~variants (OTC, OTC-Strict, Exp) for all datasets. Warm-start training is done for each individually with $l_{max}=1$.

\textbf{Results: Training with oracle helpers fails to yield selective help-seeking.}
We find that each training run converged to always asking for help within the first 50 training steps, even with stricter penalties. Note that the optimal policy should display selective help-seeking (answer parametrically for known questions) to maximize reward. However, we do not observe this in practice, as always seeking help is an easy strategy to discover. For OTC and Exponential Decay, it is given non-zero rewards for all inputs. For OTC-Strict, it is given a positive reward for each question without correct parametric answers, which is be common early in training. As such, the noisiness of the retrieval model is crucial to extract selective help-seeking in a manner aligned with its parametric knowledge. We require RL algorithms with better exploration to succeed in this oracle setting.

\begin{table}[t]
\centering
\small
\caption{Out-of-distribution accuracy (w/ and w/o search) and abstention classification results for baseline OTC, best \methodname, and best abstention models trained on HotPotQA . 
}\vspace{-2mm}
\setlength{\tabcolsep}{4.1pt}
\begin{tabular}{l|cccc}
\toprule
\textbf{Method}
& Acc$\uparrow$ & Acc w/ tool$\uparrow$ & $\text{Abs}(0)\uparrow$ & Delta$\uparrow$ \\
\midrule
\multicolumn{5}{l}{\textbf{Natural Questions}}  \\ \midrule
\textsc{otc} & 2.09 & 54.35 & 99.04 & 8.32  \\ 
\methodname~w/ \sc{otc-st}  & $18.25$ & 51.24 & 79.95 & 51.63 
 \\
DPO  & 24.4 & - & 77.38 & 68.23   \\ \midrule
\multicolumn{5}{l}{\textbf{TriviaQA}}  \\ \midrule
\textsc{otc}& 4.07 & 71.43 & 96.95 & 7.11 \\
\methodname~w/ \sc{otc-st}& 30.52 & 67.61 & 77.53 & 51.55 \\
DPO & 41.6 & - & 71.71 & 66.23 \\
\bottomrule
\end{tabular}%
\label{tab:ood_hotpotqa} \vspace{-4mm}
\end{table}


\subsection{Out-of-Distribution Performance}
\vspace{-2mm}

Finally, we evaluate our trained models out-of-distribution. Due to space, we restrict our analysis to the OTC baseline, and our best performing \methodname~variant w/ OTC-Strict and the best abstention baseline (DPO) trained on HotPotQA. We evaluate generalization to other training datasets and an additional single-hop dataset TriviaQA~\citep{joshi-etal-2017-triviaqa}.

\textbf{Results:} Table \ref{tab:ood_hotpotqa} reports our results (NaturalQA and 2Wiki models are in Appendix~\ref{sec:ood_appendix}). \methodname~generalizes better than the OTC (higher Accuracy and Delta values), which abstains on nearly all questions out-of-distribution. \methodname~also reports better Abs(0) performance than DPO, but lower Delta. We attribute this to \methodname~generalizing more conservatively out-of-domain. With 2Wiki, which only contains multi-hop questions, \methodname~generalizes well to HotPotQA but performs poorer on single-hop datasets.  We argue that, for out-of-distribution generalization, abstention and invoking search tools is the more ideal decision. With search enabled, our HotPotQA-trained \methodname~model attains $26.43\%$  higher accuracy than DPO, which is limited to abstention.

\section{Related Work}\vspace{-2mm}
\paragraph{Abstention and Verbalized Uncertainty} As LLM adoption increases, it is critical that models recognize their capability boundaries and faithfully report their uncertainty. Past work has explored this problem from many directions, developing techniques for hallucination detection~\citep{du2024haloscope, chen2024inside}, abstention~\citep{yang2024alignment, cheng2024can} and calibration~\citep{kapoor2024large}, with methods ranging from model prompting~\citep{feng-etal-2024-dont} and hidden state probing~\citep{du2024haloscope, chen2024inside} to training of the model itself~\citep{kadavath2022languagemodelsmostlyknow}. We focus on methods for abstention that train a model that intrinsically recognizes its knowledge boundaries. 

Past work predominantly relies on pipelined approaches that first estimate a model's knowledge boundaries and then use this information either to construct datasets to be used for SFT~\citep{yang2024alignment, zhang-etal-2024-r} and DPO training~\citep{cheng2024can}, or to train model-specific reward functions for RLHF~\citep{xu2024rejection}. Some work additionally seeks to align models to explicitly verbalize their uncertainty, either by explicitly constructing SFT data that summarizes uncertainty over multiple samples~\citep{xu-etal-2024-sayself} or through structured, multi-agent interaction scenarios~\citep{stengel-eskin2024lacie, eisenstein2025dont}. The work closest to ours is the Collaborative Self-Play approach of \citet{eisenstein2025dont}, which likewise investigates learning knowledge boundaries in conjunction with tool use. However, their approach relies on a specialized multi-agent scenario where each agent has a constrained set of actions and is restricted to single-hop question-answering settings. In contrast, \methodname~demonstrates that models can effectively learn their knowledge boundaries through the mere optimization of an efficient tool-use reward, without requiring any task-specific multi-agent scaffolding or assumptions on task complexity.

\paragraph{Selective RAG} There has separately been much attention on developing methods for determining when to search or continue searching in the context of retrieval augmented generation (RAG) approaches. Such methods are predominantly specialized towards the RAG setting, with methods influencing both when search is performed and the content of the search queries themselves. Approaches commonly involve estimating uncertainty, be it through operations on hidden model states~\citep{yao-etal-2025-seakr, baek-etal-2025-probing}, self-consistency over samples~\citep{ding2024retrieve} or output probabilities~\citep{jiang-etal-2023-active, su-etal-2024-dragin}. While we focus on knowledge-intensive queries, our approach is task-agnostic and only involves end-to-end reinforcement learning optimization with an efficiency reward.

\paragraph{Augmenting LLMs with Tool-Use} There has long been interest in leveraging tool-use to augment LLM capabilities~\citep{schick2023toolformer, yao2023react}, with post-training pipelines for foundation models~\citep{yang2025qwen3, team2025kimi} increasingly featuring dedicated training for tool-use. We build on top of recent work training LLMs to use search tools with RL~\citep{jin2025searchr}, particularly on top of the OTC reward formulation of~\citet{wang2025acting}. While OTC provides us one of our reward formulations,~\citet{wang2025acting} exclusively analyze gains in efficiency without reference to parametric knowledge and, furthermore, fail to answer parametrically on their experiments involving Natural Questions. We instead propose \methodname~as an alternative framework for abstention, provide analyses with varying reward formulations, consistently produce a mixture between parametric answers and search, and provide explicit comparisons against abstention methods.

\section{Conclusion}\vspace{-2mm}
We propose \methodname, a novel framework that trains LLMs  for selective help-seeking, and readily extracting abstention behaviors. \methodname~trains models for two capabilities at the cost of one -- models learn how to use search tools and synthesize information, and distinguish between answerable/unanswerable questions. Our results on 3 short-form knowledge-intensive datasets show that \methodname~outperforms previous efficient search baselines on overall accuracy when allowed searches and also demonstrates strong abstention behaviors, analogous to specialized abstention methods. 


\section{Acknowledgments}

This project was partially supported by NSF grant IIS-2433072, and a gift from Google. We gratefully acknowledge use of the research computing resources of the Empire AI Consortium, Inc, with support from Empire State Development of the State of New York, the Simons Foundation, and the Secunda Family Foundation. We thank the Cornell NLP group for helpful discussions and comments.

\bibliography{iclr2026_conference}
\bibliographystyle{icml2026}

\appendix
\onecolumn
\section{Theoretical Analysis}\label{sec:theoretical_analysis}
\subsection{Preliminaries}
In this section, we provide a theoretical analysis of the search behavior the optimal policy for the proxy objective of Equation \ref{eq:proxy_objective} will display for a given question. Specifically, we will formally demonstrate that the optimal policy must produce a parametric answer to a question $q$ if and only if the maximum achievable expected accuracy when answering $q$ parametrically is greater than or equal to the maximum achievable expected reward when performing searches.

Before starting the analysis proper, we will first make two assumptions: 
\begin{enumerate}[leftmargin=*] 
    \item[-] \textbf{Assumption 1:} We set the KL penalty weight $\beta = 0.0$ to simplify the proxy objective and focus solely on the search behavior of an optimal policy maximizing reward. 
    \item[-] \textbf{Assumption 2:} We assume that the optimal policy cannot achieve perfect expected accuracy (i.e., average accuracy across multiple samples), for all questions when answering parametrically. In other words, the maximum achievable expected accuracy is bounded for some questions. This follows by definition from our training setup, where an LLM cannot acquire new knowledge beyond the base model's knowledge boundaries during RL and from prior work, which suggests that standard finetuning approaches on new knowledge risk an increase in hallucinations rather than incorporation of knowledge~\citep{lin2024flame, gekhman-etal-2024-fine}.
\end{enumerate}

We then make three definitions to be used in our proofs:
\begin{enumerate}[leftmargin=*] 
    \item[-] \textbf{Definition 1:} Let $N_s(\tau)$ be the number of search calls made by a trajectory $\tau$.
    \item[-] \textbf{Definition 2:} Let $r_{acc}(y, \tau) \in \{0, 1\}$ and $r_{help}(q, \tau)$ be defined such that $r_{acc}(y, \tau)\cdot r_{help}(q, \tau) = 1$ if $r_{acc}(y, \tau) = 1$ and $N_s(\tau) = 0$. This follows from the definitions set in Section \ref{sec:problem_defn}.
    \item[-] \textbf{Definition 3:} Consider a question-answer pair $(q, y)$. Then, we define: $$R^*_i(q, y) = \max_{\theta} E_{\tau \sim \pi_\theta(\cdot |q, H)}[r_{acc}(y, \tau) \cdot r_{help}(q, \tau) | N_s(\tau) = i]$$
    as the maximum achievable expected reward by a policy when performing $i$ searches.
\end{enumerate}

\subsection{Proof}
\paragraph{Lemma 1:} For a given question-answer pair $(q, y)$, the optimal policy $\pi_{\theta^*}$ must answer the question $q$ parametrically (without any searches) if and only if $R^*_0(q,y) > R^*_i(q,y)$ for all $i > 0$.

\paragraph{Proof of Lemma 1:} We firstly note that the expected reward for any arbitrary policy $\theta$ given a question-answer pair $(q, y)$ can be written as a weighted sum of its expected reward when answering question $q$ with different search counts. We show this below:
\begin{align*}
E_{\tau \sim \pi_{\theta}(\cdot | q, H)}&[r_{acc}(y, \tau) \cdot r_{help}(q, \tau)] = \Sigma_\tau \left [ \pi_{\theta}(\tau | q, H) \cdot r_{acc}(y, \tau) \cdot r_{help}(q, \tau) \right ]\\
&= \Sigma^\infty_{i=0} \Sigma_{\tau | N_s(\tau) = i} \left [ \pi_{\theta}(\tau | q, H) \cdot r_{acc}(y, \tau) \cdot r_{help}(q, \tau) \right ] \\
&= \Sigma^\infty_{i=0} \left [ P(N_s(\tau) = i | q; \theta) \cdot \Sigma_{\tau | N_s(\tau) = i} \left [ \frac{\pi_{\theta}(\tau | q, H) }{ P(N_s(\tau) = i | q; \theta) } \cdot r_{acc}(y, \tau) \cdot r_{help}(q, \tau) \right ] \right ] \\
&= \Sigma^\infty_{i=0} \left [ P(N_s(\tau) = i | q; \theta) \cdot E_{\tau \sim \pi_{\theta}(\cdot | q, H)} \left [r_{acc}(y, \tau)\cdot r_{help}(q, \tau) | N_s(\tau) = i \right ] \right ],
\end{align*}
where $P(N_s(\tau) = i | q; \theta)$ is the probability that a policy parameterized by $\theta$ will produce a trajectory $\tau$ with $N_s(\tau) = i$ given question $q$.

\textbf{Given this, we will first prove the forward direction.} Let $\theta^*$ be the parameters optimizing the objective and let $(q,y)$ be an arbitrary question-answer pair. Assume that $\pi_{\theta^*}$ must answer question $q$ parametrically. This firstly implies that $P(N_s(\tau) = 0|q;\theta^*) = 1$ and, due to the optimality of $\theta^*$, that
\begin{align*}
    E_{\tau \sim \pi_{\theta^*}(\cdot |q, H)}[r_{acc}(y, \tau) \cdot r_{help}(q, \tau) | N_s(\tau) = 0] &= \max_{\theta} E_{\tau \sim \pi_\theta(\cdot |q, H)}[r_{acc}(y, \tau) \cdot r_{help}(q, \tau) | N_s(\tau) = 0] \\
    &= R^*_0(q,y).
\end{align*}
Furthermore, we note that the fact that $\theta^*$ must answer $q$ parametrically implies that
\begin{equation*}
    R^*_0(q,y) = E_{\tau \sim \pi_{\theta^*}(\cdot |q, H)}[r_{acc}(y, \tau) \cdot r_{help}(q, \tau) | N_s(\tau) = 0] > R^*_i(q,y)
\end{equation*}
for all $i > 0$, where $R^*_i(q, y)$ is the maximum attainable expected reward when using $i$ searches. Assume for the sake of contradiction this were not true and that there existed some $i$ such that 
\begin{equation*}
    R^*_0(q, y) = E_{\tau \sim \pi_{\theta^*}(\cdot |q, H)}[r_{acc}(y, \tau) \cdot r_{help}(q, \tau) | N_s(\tau) = 0] \le R^*_i(q,y).
\end{equation*}
Let $\hat{\theta}$ be a policy that sets $P(N_s(\tau) = i|q; \hat{\theta}) = 1$ and displays the same behavior as $\theta^*$ all other questions. In the first case, $R^*_0(q, y) = R^*_i(q, y)$. Here, $\hat{\theta}$ would have the same expected reward as $\theta^*$ and be another optimal policy. However, this contradicts our assumption that the optimal policy \textit{must} answer $q$ parametrically. In the other case, $R^*_0(q, y) < R^*_i(q, y)$. In this instance, $\hat{\theta}$ achieves a greater expected reward than $\theta^*$, which contradicts the assumption that $\theta^*$ is optimal. Hence, $R^*_0(q, y) > R^*_i(q, y)$ for all $i > 0$, which proves the forward direction.

\textbf{We will next prove the backwards direction.} Let $\theta^*$ be the parameters optimizing the objective and assume that $R^*_0(q, y) > R^*_i(q, y)$. We claim that $\theta^*$ must answer $q$ parametrically. Let $C$ be an arbitrary set of non-negative integers containing at least one positive integer $i > 0$ and assume for the sake of contradiction that $\theta^*$ set $P(N_s(\tau) = i|q; \theta^*) > 0$ for all $i \in C$ and $\sum_{i \in C}P(N_s(\tau) = i|q; \theta^*) = 1$. Then, it firstly follows that:
\begin{align*}
    E_{\tau \sim \pi_{\theta^*}(\cdot | q, H)}[r_{acc}(y, \tau) \cdot r_{help}(q, \tau)] &= \sum_{i \in C}P(N_s(\tau) = i|q; \theta^*)E_{\tau \sim \pi_{\theta^*}(\cdot | q, H)}[r_{acc}(y, \tau) \cdot r_{help}(q, \tau)|N_s(\tau) = i] \\
    &= \sum_{i \in C}P(N_s(\tau) = i|q; \theta^*)R^*_i(q, y),
\end{align*}
due to our assumption that $\theta^*$ is optimal. However, due to our assumption that $R^*_0(q, y) > R^*_i(q, y)$ for all $i > 0$, we also have that
\begin{equation*}
    \sum_{i \in C}P(N_s(\tau) = i|q; \theta^*)R^*_i(q, y) < \sum_{i \in C}P(N_s(\tau) = i|q; \theta^*)R^*_0(q, y) = R^*_0(q, y).
\end{equation*}
Given this, we can construct an alternative policy $\hat{\theta}$ that sets $P(N_s(\tau) = 0|q;\hat{\theta}) = 1$ and behaves the same as $\theta^*$ on all other questions. This policy $\hat{\theta}$ will have a larger expected reward than $\theta^*$, which contradicts our assumption that $\theta^*$ is optimal. Hence, $\theta^*$ cannot set $P(N_s(\tau) = i|q;\theta^*) > 0$ for any $i > 0$ and \textit{must} answer $q$ parametrically. This proves the reverse direction. As a result, our claim holds.

\paragraph{Lemma 2:} Let $(q, y)$ be an arbitrary question-answer pair. Then, the maximum achievable expected accuracy when answering $q$ parametrically is equal to $R^*_0(q, y)$.

\paragraph{Proof of Lemma 2:} Consider an arbitrary trajectory $\tau$ such that $N_s(\tau) = 0$. We claim that the accuracy of $\tau$ is equal to the reward of $\tau$. In the first possibility, $r_{acc}(y, \tau) = 1$. Then, that the reward $r_{acc}(y, \tau) \cdot r_{help}(q, \tau) = 1$ follows from Definition 2. In the second possibility, $r_{acc}(y, \tau) = 0$. Then, the reward $$r_{acc}(y, \tau) \cdot r_{help}(q, \tau) = 0 \cdot r_{help}(q, \tau) = 0.$$
Hence, the accuracy and reward of $\tau$ are equal when $N_s(\tau) = 0$. This further implies that the maximum achievable expected accuracy is equal to the maximum achievable expected reward when answer $q$ parametrically. This latter quantity, by Definition 3, is $R^*_0(q, y)$. The lemma is therefore correct.

\paragraph{Theorem:} For a given question-answer pair $(q, y)$, the optimal policy $\pi_{\theta^*}$ must answer the question $q$ parametrically (without any searches) if and only if the maximum achievable expected accuracy when answering $q$ parametrically is greater than or equal to the maximum achievable expected reward when performing searches.

\paragraph{Proof of Theorem:} Consider an arbitrary question-answer pair $(q, y)$ and let $\theta^*$ be the optimal policy. Then, Lemma 1 implies that $\theta^*$ must answer the question $q$ parametrically if and only if $R^*_0(q, y) > \max_{i > 0}R^*_i(q, y)$. Furthermore, by Lemma 2, we know that the maximum achievable expected accuracy when answering $q$ is equal to $R^*_0(q, y)$. These two lemmas together imply that the optimal policy $\pi_{\theta^*}$ must answer the question $q$ parametrically if and only if the maximum achievable expected accuracy when answering $q$ parametrically is greater than or equal to the maximum achievable expected reward when performing searches. The theorem holds.

\section{Search Tool Use}\label{sec:search_tool_training_appendix}
In this section, we provide details for GRPO and warm-start training and describe the datasets used for training and evaluation.

\subsection{GRPO Training}
\label{sec:app_grpo_training}
We use the GRPO implementation of the veRL library~\citep{10.1145/3689031.3696075} for all RL training. 

\paragraph{Training hyperparameters} For general training hyperparameters, we set the learning rate to $10^{-6}$ without any warmup or decay and use a gradient clipping norm of $1.0$. For policy optimization, we set $\epsilon=0.2$, entropy coefficient to $0.001$, batch size to $64$, group size $G=16$ and perform $1$ gradient step per rollout. In early hyperparameter tuning experiments, we observed setting $\beta=0$ to improve performance, with the associated benefit of freeing the memory used for the reference model. In doing so, we follow other follow-up work on GRPO~\citep{liu2025understanding}.

We perform training for $400$ steps and evaluate the model on the task's validation set every $25$ steps. We restrict the use of LLM judges only to the test set and use exact match to estimate accuracy for training and validation. We pick the checkpoint to evaluate using validation tool productivity performance.

\paragraph{Retrieval details} We use the retrieval server implementation provided by Search-R1~\citep{jin2025searchr} for retrieval. We further follow Search-R1 in masking out tokens from retrieved documents when computing losses. We use the E5 retriever~\citep{wang2022text} with $3$ documents returned per query. We enclose each returned query in-between $<$document$>$ tags.

\paragraph{Inference hyperparameters} We perform inference with a temperature of $1.0$ during both training and test, and do not use either top-$p$ or top-$k$ sampling. The maximum output length for an individual generation step is $512$ tokens and we set the maximum overall output length (with retrieved documents added) to $6144$. We truncate outputs exceeding the maximum output length. 

\paragraph{Input prompts} We use the prompt shown in Figure \ref{fig:baseline_prompt} for tool-use training. This is based on the prompt used by \citet{wang2025acting}. For R1 training, on the other hand, we use the prompt shown in Figure \ref{fig:r1_prompt}. This is identical to the R1 prompt used in Search-R1.

\subsection{Inference Algorithm} Inference is done according to the procedure detailed in Algorithm \ref{alg:tool_use_inference}. Note that this inference procedure during RL training and evaluation is distinct from the structured inference procedure used in warm-start data generation (as described in Algorithm \ref{alg:warm_start}). If a model exceeds the maximum number of allowed searches and still attempts a search, it is given a warning message instead. We observed that this did not occur for runs featuring the efficiency reward. We set the maximum number of searches in our Search-R1 experiments to $3$ due to compute and memory concerns. Finally, we do not manually append a course-correction message upon failure to generate a properly formatted search or answer tag, as this is a task-specific addition and must be defined for each tool individually.

\begin{algorithm}[h]
  \captionof{algorithm}{Inference with Multi-Turn Search Tool Calls}\label{alg:tool_use_inference}
\begin{algorithmic}
\REQUIRE Question $q$, language model $\pi_\theta$, retriever $H$
\HYPERPARAMETER Maximum search budget $L$
\ENSURE Trajectory $\tau$
\STATE Initialize trajectory $\tau \gets \emptyset$
\STATE Initialize action count $l \gets 0$
\WHILE{$l \le L+2$}
    \STATE Generate action $a_l \sim \pi_\theta(\cdot | q, \tau; H)$ until [$<$/search$>$, $<$/answer$>$, $<$eos$>$]
    \STATE Append $a_l$ to trajectory $\tau \gets \tau + a_l$
    \IF{$<$search$>$ $<$/search$>$ detected in $a_l$ and $l < L$}
        \STATE Extract search query $s_l$
        \STATE Retrieve top-$k$ documents $o_l \sim H(s)$
        \STATE Append documents to trajectory $\tau \gets \tau + o_l$
    \ELSIF{$<$search$>$ $<$/search$>$ detected in $a_l$}
        \STATE Construct warning message $m = $ $<$warning$>$ SEARCH LIMIT REACHED $<$/warning$>$
        \STATE Append $m$ to trajectory $\tau \gets \tau + m$
    \ELSIF{$<$answer$>$ $<$/answer$>$ detected in $a_l$ or $<$eos$>$ detected in $a_l$}
        \STATE \RETURN Final generated response $\tau$
    \ENDIF
    \STATE Increment $l \gets l + 1$
\ENDWHILE
\STATE \RETURN $\tau$
\end{algorithmic}
\end{algorithm}

\begin{table*}[t]
\centering
\small
\caption{Abstention classification results for the warm-start initializations. We report Abs(0), i.e. \% abstention for unanswerable questions (higher better), and the delta between the \% abstention between unanswerable and answerable questions.}
{%
\begin{tabular}{l|cc|cc|}
\toprule
\multirow{2}{*}{\textbf{Method}} & \multicolumn{2}{c|}{\textbf{Natural Questions}} & \multicolumn{2}{c|}{\textbf{HotPotQA}} \\
\cmidrule(lr){2-3} \cmidrule(lr){4-5}
& $\text{Abs}(0)\uparrow$ & Delta$\uparrow$ & $\text{Abs}(0)\uparrow$ & Delta$\uparrow$   \\
\midrule
Warm-Start Initialization  & 66.17 & 1.55 & 68.64 & 7.69  \\
\bottomrule
\end{tabular}%
} 

\label{tab:warm_start_abstention_curve}
\end{table*}

\subsection{Warm-Start}
\label{sec:app_warm_start}
\paragraph{Warm-Start Implementation Details} We follow the procedure outlined in Algorithm \ref{alg:warm_start} to construct the warm-start data. We use the Qwen2.5-32B base model as our generator, as it is better capable of following instructions off-the-shelf, but has not undergone alignment for abstention unlike instruct models. Nonetheless, to ensure that the base model generates properly formatted outputs, we sample $4$ candidate outputs for each action and discard the output if it contains unrelated tags or does not add the action ending tag. For think and search actions, we choose a random output. For answer actions, we preferentially choose correct outputs.

Evaluation of trajectories is done with an LLM judge, in this case Qwen2.5-72B-Instruct~\citep{qwen2025qwen25technicalreport}. We follow the same procedure we use to evaluate abstention model outputs, described in Section \ref{sec:question_accuracy_estimation}. If a trajectory is deemed correct, we swap its generated answer with the ground-truth answer for the target dataset to align answers with the dataset format, as we use exact match as the reward.

For a given question $q$, if we sample $l=0$ as the target number of actions, we use the prompt used for R1 training (Figure \ref{fig:r1_prompt}) to prevent the model from searching. Otherwise, we use the prompt described in Figure \ref{fig:decomposition_prompt}.

\begin{algorithm}[t]
\caption{Warm-Start Trajectory Construction}\label{alg:warm_start}
\begin{algorithmic}
\REQUIRE Datapoint $(q, a^*)$, generator $G$, retriever $H$, maximum searches $l_{max}$, num samples $N$
\ENSURE Datapoint $(q, \tau)$ for SFT
\STATE Sample random number of searches $l \sim \{0, \ldots, l_{max}\}$
\STATE Define $\text{seq} \gets $ [\texttt{\textbf{\small think}}, \texttt{\small\textbf{search}}] $ \times\; l \;+ \;$ [\texttt{\small\textbf{think}}, \texttt{\small \textbf{answer}}]
\FOR{$i=1 \to N$}
\STATE Initialize current trajectory $\tau_i \gets \emptyset$
 \FOR{action in $\text{seq}$}
     \STATE Append action start tag $\tau_i \gets \tau_i ~+ <\text{action}>$
     \STATE Generate action $a \sim G(\cdot | q, \tau_i)$ until $<$/action$>$
     \STATE Append action $a$ to trajectory $\tau_i \gets \tau_i + a$
     \IF{action $=$ search}
         \STATE Retrieve top-$k$ documents $o \sim H(a)$ 
         \STATE Append $o$ to trajectory  $\tau_i \gets \tau_i + o$
     \ENDIF
 \ENDFOR
\ENDFOR
\STATE Set $\tau$ to a random correct $\tau_i$ if any, else $\tau_i$ with shortest answer. 
\STATE \RETURN $\tau$
\end{algorithmic} 
\end{algorithm} 

\paragraph{Training Details} We use Huggingface TRL's SFTTrainer to perform training~\citep{vonwerra2022trl}. We use the hyperparameters used by~\citet{muennighoff2025s1simpletesttimescaling} for performing SFT on reasoning data. Specifically, we use a learning rate of $10^{-5}$, weight decay of $10^{-4}$, Adam $\beta_1=0.9, \beta_2=0.95$ and gradient clipping norm of $1$. We use a linear learning rate scheduler warmed-up for $5\%$ of training steps and decayed to $0$ throughout training. We train for $5$ epochs with an effective batch size of $16$. As in RL training, tokens corresponding to retrieved documents are masked out from the loss.

\paragraph{Lack of alignment with parametric knowledge} In Table \ref{tab:warm_start_abstention_curve}, we report our warm-start initializations' performance in terms of the Abs(0) and Delta metrics (as defined in Section \ref{sec:setup_abstention}). On both Natural Questions and HotPotQA, the warm-start initialization has miniscule Delta values of $1.55$ and $7.69$, indicating that the model does not behave differently for unanswerable and answerable questions. Furthermore, as we set $l_{max}=2$ and choose the target number of searches in warm-start data randomly, two thirds of the data has search (and, therefore, abstention) behavior. This explains the Abs(0) values near $66\%$.

\subsection{Datasets}
\label{sec:app_datasets}
We run training experiments on three knowledge-intensive datasets -- the single-hop dataset Natural Questions~\citep{kwiatkowski2019natural}, and multi-hop datasets HotPotQA~\citep{yang-etal-2018-hotpotqa} and 2WikiMultiHopQA~\citep{ho-etal-2020-constructing}. Table \ref{tab:dataset_answerability_distribution} displays the distribution of question answerability for each dataset. Specifically, each column represents the proportion of test set questions where Qwen2.5-3B-Base achieves an average accuracy of $i$, estimated using the procedure outlined in Appendix \ref{sec:question_accuracy_estimation}. As seen in the table, the distributions are very similar for Natural Questions and HotPotQA, with $44.72\%$ and $50.11\%$ of questions having an average accuracy of $0$ and $11.26\%$ and $7.92\%$ of questions having an average accuracy of $1$. 2WikiMultiHopQA is more challenging than both, with $77.26\%$ of questions having an average accuracy of $0$ and only $0.85\%$ of questions having average accuracy of $1$. 

We additionally use the single-hop TriviaQA dataset as part of our out-of-distribution evaluations. For Natural Questions, we use the official splits for training, validation and test. For HotPotQA, 2WikiMultiHopQA and TriviaQA, the official test splits do not contain answers. As a result, we use their official development/validation sets for the purpose of test and construct our own validation sets by sub-sampling from the training set with a 90/10 split. Additionally, as noted in the main text, we filter out the ``comparison'' and ``bridge-comparison'' questions from 2WikiMultiHopQA, as these questions are each binary choice questions with heavily skewed answer distributions, causing models to reward hack in practice.

\begin{table*}[t]
\centering
\small
\caption{Proportion of test set questions where Qwen2.5-3B-Base achieves an average accuracy of $i$ for each $i \in \{0, 0.1, \ldots, 0.9, 1\}$ for Natural Questions, HotPotQA and 2WikiMultiHopQA. We find that Natural Questions and HotPotQA have similar distributions of average accuracy, with 2WikiMultiHopQA being harder than both}
\setlength{\tabcolsep}{2.5pt}
\begin{tabular}{l|ccccccccccc|}
\toprule
\multirow{2}{*}{\textbf{Method}} & \multicolumn{11}{c|}{\textbf{Percentage of Questions with an Average Accuracy of $i$}} \\
\cmidrule(lr){2-12}
& 0 & 0.1 & 0.2 & 0.3 & 0.4 & 0.5 & 0.6 & 0.7 & 0.8 & 0.9 & 1 \\
\midrule
Natural Question  & 44.72 & 9.62 & 6.42 & 5.06 & 4.04 & 3.07 & 3.65 & 4.04 & 3.73 & 4.40 & 11.26 \\
HotPotQA & 50.11 & 9.41 & 5.91 & 4.75 & 3.86 & 3.94 & 3.54 & 3.47 & 3.71 & 3.37 & 7.92 \\
2WikiMultiHopQA & 77.26 & 8.49 & 3.77 & 2.46 & 1.97 & 1.33 & 1.22 & 0.96 & 0.85 & 0.83 & 0.85 \\
\bottomrule
\end{tabular}%
\vspace{-1mm}
\label{tab:dataset_answerability_distribution}
\end{table*} 
\begin{figure*}[t!]
    \centering
    \fbox{
        \parbox{\dimexpr\textwidth-2\fboxsep-2\fboxrule\relax}{\small
        \textbf{Input Prompt}: \\
Answer the given question. You should first have a reasoning process in mind and then provides the answer. Show your reasoning in $<$think$>$ $<$/think$>$ tags and return the final answer in $<$answer$>$ $<$/answer$>$ tags, for example $<$answer$>$ Beijing $<$/answer$>$. Question: $<$question$>$
        }
    }
    \caption{The input prompt used during R1 training experiments. The final $<$question$>$ is replaced by the input question.}
    \label{fig:r1_prompt}
\end{figure*}
\begin{figure*}[t!]
    \centering
    \fbox{
        \parbox{\dimexpr\textwidth-2\fboxsep-2\fboxrule\relax}{\small
        \textbf{Input Prompt}: \\
Answer the given question. You must conduct reasoning between $<$think$>$ and $<$/think$>$ every time you get new information. After reasoning, if you find you lack some knowledge, you can call a search engine by $<$search$>$ query $<$/search$>$ and it will return the top searched results between $<$document$>$ and $<$/document$>$. You need to make every search call count and gain helpful results. If you find no further external knowledge is needed, you can directly provide the answer inside $<$answer$>$ and $<$/answer$>$, without detailed illustrations. For example, $<$answer$>$ Beijing $<$/answer$>$. Question: $<$question$>$
        }
    }
    \caption{The input prompt used during search tool use experiments. The final $<$question$>$ is replaced by the input question.}
    \label{fig:baseline_prompt}
\end{figure*}
\begin{figure*}[t!]
    \centering
    \fbox{
        \parbox{\dimexpr\textwidth-2\fboxsep-2\fboxrule\relax}{\small
        \textbf{Input Prompt}: \\
Answer the given question. You must conduct reasoning between $<$think$>$ and $<$/think$>$ every time you get new information. After reasoning, if you find you lack some knowledge, you can ask a question to a search engine by $<$search$>$ query $<$/search$>$ and it will return the top searched results between $<$document$>$ and $<$/document$>$. A search query should be an atomic question asking about one, single piece of information. \\

Example 1:

Question: ``Who was born first, Clint Eastwood or Harrison Ford?''

Valid Queries: ``$<$search$>$Clint Eastwood birth date$<$/search$>$'' and ``$<$search$>$Harrison Ford birth date$<$/search$>$''.

The query ``$<$search$>$Clint Eastwood and Harrison Ford birth date$<$/search$>$'' is invalid. 

The query ``\\ $<$search$>$ \\ Clint Eastwood birth date \\ Harrison Ford birth date \\ $<$/search$>$'' \\ is also invalid. Do not pack in multiple questions into one query. Each query should be completely independent. \\

Example 2:

Question: ``Which is a genus of palms, Zinnia or Butia?''

Valid Queries: ``$<$search$>$Zinnia genus classification$<$/search$>$'' and ``$<$search$>$Butia genus classification$<$/search$>$''. \\

Example 3:

Question: ``When did the country where Piltene is located become part of the USSR?''

Initial Query: ``$<$search$>$Piltene location$<$/search$>$'' \\

In each of these examples, you should conduct a search only if you lack the relevant information. Remember, you should decompose questions in your search queries and conduct searches for each atomic question separately. You need to make every search call count and gain helpful results. If you find no further external knowledge is needed, you can directly provide the answer inside $<$answer$>$ and $<$/answer$>$, without detailed illustrations. For example, 
$<$answer$>$ Beijing $<$/answer$>$. \\

Question: $<$question$>$
        }
    }
    \caption{The input prompt used when generating tool-use trajectories during warm-start data generation. The final $<$question$>$ is replaced by the input question.}
    \label{fig:decomposition_prompt}
\end{figure*}

\section{Abstention Experiment Details}
\label{sec:app_abstention_details}
In this section, we first detail the pipeline for estimating the average accuracy the base model achieves on each question. This is used to determine both answerability boundaries for abstention training as well as compute abstention classification metrics. We then describe training and inference for our abstention methods.

\subsection{Question Accuracy Estimation}\label{sec:question_accuracy_estimation}
We follow the pipeline used by ~\citet{yang2024alignment} to estimate the average accuracies. For a given question $q$, we sample $10$ responses $\{\hat{y_i}\}^{10}_{i=1}$ from the untrained model. As all of our experiments are conducted with base models, we perform few-shot prompting. Specifically, for each dataset, we collect correct responses sampled from DeepSeek-V3.1 to $5$ questions sampled from the training set and use these as our few-shot examples. For this component, we perform inference with DeepSeek-V3.1 using a temperature of $1$ and top-$p$ of $0.8$. We likewise perform sampling with Qwen2.5-3B with a temperature of $1$, top-$p$ of $0.8$ and top-$k$ of $50$ to ensure that the base model samples strong outputs and gives a good estimate of knowledge boundaries.

To assess the correctness of a given answer $\hat{y}_i$, we first extract a shortform response and then evaluate the accuracy of this extracted response with an LLM judge. We use DeepSeek-V3.1 in both cases using the few-shot prompts of \citet{yang2024alignment} (shown in Figures \ref{fig:answer_extraction_prompt} and \ref{fig:answer_evaluation_prompt}), using greedy decoding for replicability.

\subsection{Trained Abstention Model Details}
For both the Alignment for Honesty~\citep{yang2024alignment} and DPO~\citep{rafailov2023direct} baselines, we use the exact same training datapoints that \methodname~was trained on. Furthermore, we perform the exact same number of gradient steps to ensure a fair comparison.

For the Alignment for Honesty variants, we use Huggingface TRL's SFTTrainer~\citep{vonwerra2022trl}. We use a learning rate of $10^{-5}$, weight decay of $10^{-4}$, Adam $\beta_1=0.9, \beta_2=0.95$ and gradient clipping norm of $1$. We use a linear learning rate scheduler warmed-up for $5\%$ of training steps and decayed to $0$ throughout training. For the ``Absolute'' variant of Alignment for Honesty, we use an effective batch size of $64$. For the ``Multisample'' variant, we use an effective batch size of $640$ to achieve the same number of gradient steps, as it constructs a datapoint for each question-answer pair sampled during average accuracy estimation.

For the DPO baseline, we use Huggingface TRL's DPOTrainer. While we take inspiration from \citet{cheng2024can} in constructing the preference dataset, we do not use their two-stage approach featuring an initial SFT stage followed by a DPO stage. Instead, we find that doing DPO training with SFT regularization performs well~\citep{pang2024iterative} and is more comparable to our other settings. We use the same hyperparameters as in the Absolute variant of Alignment for Honesty. We set the DPO $\beta=0.1$ and the SFT loss coefficient to $1$.

Both models are trained to respond to the prompt shown in Figure \ref{fig:abstention_prompt}. We perform inference with a temperature of $1.0$, without top-$p$ or top-$k$ sampling, as is done for our \methodname~models.

\subsection{Few-Shot Abstention Prompting Details}
For few-shot prompting, we likewise use the prompt shown in Figure \ref{fig:abstention_prompt}. As mentioned in Section \ref{sec:experiments}, we average performance over $4$ samples. In the case of the few-shot abstention prompt, we use a separate few-shot prompt for each sample. Two of the few-shot prompts feature $3$ abstentions on unanswerable questions and $2$ answers on always answerable questions. The other two feature $3$ answers on always answerable questions and $2$ abstentions on unanswerable ones. The answers themselves are sampled from DeepSeek-V3.1.

\begin{figure*}[t!]
    \centering
    \fbox{
        \parbox{\dimexpr\textwidth-2\fboxsep-2\fboxrule\relax}{\small
        \textbf{Input Prompt}: \\
Answer the given question. If you are not confident that your answer will be correct, you should abstain from answering by using the phrase 'I am afraid I cannot help you as I do not know the answer to this question.' Question: $<$question$>$
        }
    }
    \caption{The input prompt used during ternary abstention experiments. The final $<$question$>$ is replaced by the input question.}
    \label{fig:ternary_prompt}
\end{figure*}

\subsection{Ternary Abstention Baseline Details}
\begin{figure*}
     \centering
     \vspace{4mm}
     \includegraphics[scale=0.3,trim=0mm 0mm 0mm 15mm]{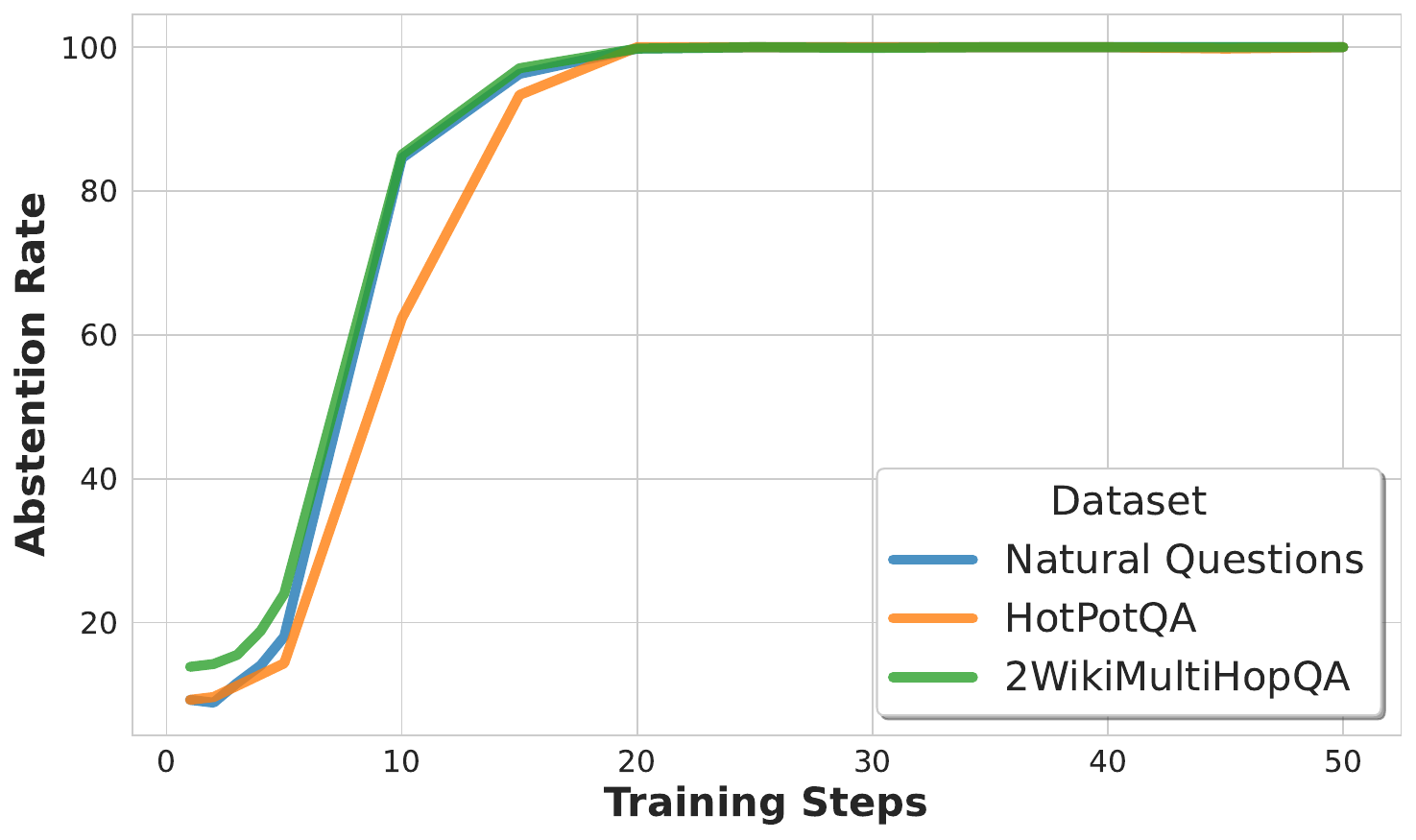}
     \caption{Abstention rate at different training steps when trained with a ternary reward for abstention. Models converge to always abstaining within 25 steps for all datasets.}
     \label{fig:ternary_abstention_plot} 
 \end{figure*}

The Ternary baseline is trained with GRPO using a ternary reward that rewards correct answers with $+1$, abstentions with $0$ and incorrect answers with $-1$. We assess correctness using exact match and assess abstention heuristically by checking whether a string from the list [\texttt{i am afraid}, \texttt{i'm afraid}, \texttt{i cannot help}, \texttt{i can't help}, \texttt{do not know}, \texttt{don't know}, \texttt{abstain}, \texttt{abstention}] appears as a substring in the final answer. Training uses the prompt shown in Figure \ref{fig:ternary_prompt}. Similar to our oracle helper setting, we find that training with this ternary reward leads to models always abstaining within $25$ steps. This can be seen in Figure \ref{fig:ternary_abstention_plot}.

\subsection{Evaluating Abstention Models}
The prompt (Figure \ref{fig:answer_extraction_prompt}) used for extracting shortform answers by \citet{yang2024alignment} additionally contains few-shot examples for abstention. As a result, we first determine if a response contains an abstention using this prompt. If it does not contain an abstention, then we evaluate the extracted answer using the prompt in Figure \ref{fig:answer_evaluation_prompt}.

\begin{figure*}[t!]
    \centering
    \fbox{
        \parbox{\dimexpr\textwidth-2\fboxsep-2\fboxrule\relax}{\small
        \textbf{Input Prompt}: \\
Given a question and a piece of text, if the text does not contain an answer to the question, output ``no answer''; otherwise, extract the answer from the text. \\

Question: What was the last US state to reintroduce alcohol after prohibition? \\
Text: The last US state to reintroduce alcohol after prohibition was Mississippi. Mississippi legalized alcohol on August 17, 1933, making it the last state to do so.

Output: Mississippi 

...

Question: $<$question$>$ \\
Text: $<$model response$>$ \\ 
Output:
        }
    }
    \caption{The input prompt used to extract shortform answers from model outputs during abstention model evaluation and average accuracy estimation for questions.}
    \label{fig:answer_extraction_prompt}
\end{figure*}
\begin{figure*}[t!]
    \centering
    \fbox{
        \parbox{\dimexpr\textwidth-2\fboxsep-2\fboxrule\relax}{\small
        \textbf{Input Prompt}: \\
Please rate the consistency between the reference answer and the proposed answer on a scale of 0 to 1. A rating of 0 indicates inconsistency, while a rating of 1 indicates perfect consistency. \\

Question: In which country is the Sky Train Rail bridge? \\
Reference Answer: Canada \\
Proposed Answer: Thailand \\
Score: 0 \\
... \\
Question: $<$question$>$ \\
Reference Answer: $<$gold answer$>$ \\ 
Proposed Answer: $<$extracted answer$>$ \\
Score:
        }
    }
    \caption{The input prompt used to evaluate model answers. We follow \citet{yang2024alignment} in treating an output score higher than $0.7$ as indicating correctness.}
    \label{fig:answer_evaluation_prompt}
\end{figure*}
\begin{figure*}[t!]
    \centering
    \fbox{
        \parbox{\dimexpr\textwidth-2\fboxsep-2\fboxrule\relax}{\small
        \textbf{Input Prompt}: \\
Answer the given question. If you are not confident that your answer will be correct, you should abstain from answering by using the phrase ``I am afraid I cannot help you as I do not know the answer to this question.''

Question: $<$question$>$
        }
    }
    \caption{The input prompt used in our abstention models.}
    \label{fig:abstention_prompt}
\end{figure*}
\begin{figure*}[t!]
    \centering
    \fbox{
        \parbox{\dimexpr\textwidth-2\fboxsep-2\fboxrule\relax}{\small
        \textbf{Input Prompt}: \\
Answer the given question. You must conduct reasoning between $<$think$>$ and $<$/think$>$ every time you get new information. After reasoning, if you find you lack some knowledge, you can ask for help by $<$help$>$ I need help $<$/help$>$ and it will return the answer to the original question between $<$helper\_answer$>$ and $<$/helper\_answer$>$. You need to ask for help only when necessary. If you find no further external knowledge is needed, you can directly provide the answer inside $<$answer$>$ and $<$/answer$>$, without detailed illustrations. For example, $<$answer$>$ Beijing $<$/answer$>$. Question: $<$question$>$

        }
    }
    \caption{The input prompt used during oracle helper experiments. The final $<$question$>$ is replaced by the input question.}
    \label{fig:oracle_helper_prompt}
\end{figure*}

\section{Generalization to Other Model Scales and Families Details}
\label{sec:app_different_models}
To demonstrate that our insights regarding MASH generalize to models of different scales and families, we conduct further experiments with Qwen2.5-7B-Base and Qwen3-4B-Base respectively. We focus on the HotPotQA dataset due to its mixture of single- and multi-hop questions for these experiments. We conduct RL training under the OTC and \methodname~w/ OTC-Strict settings and further compare against each abstention baseline, with the exception of Ternary, which trivially learned to abstain on all questions for Qwen2.5-3B-Base. Due to compute limitations, we restrict these experiments to $300$ training steps, as opposed to the $400$ steps for Qwen2.5-3B-Base. We show results for the first inference setting with search allowed on Tables \ref{tab:add_models_tool_use} and \ref{tab:added_models_search_distribution}, and abstention results on Table \ref{tab:added_models_abs}.

\begin{table*}[t]
\centering
\small
\caption{Accuracy, average number of tool calls (TC) and tool productivity (TP) statistics for OTC and \methodname~w/ OTC-ST evaluated under the \textbf{inference w/  search tools} setting on HotPotQA. \methodname~w/ OTC-ST continues to outperform the OTC baseline on both Accuracy and TP, achieving a $10\%$ increase on the latter.}
\begin{tabular}{l|ccc|ccc|}
\toprule
\multirow{2}{*}{\textbf{Method}} & \multicolumn{3}{c|}{\textbf{Qwen2.5-7B-Base}} & \multicolumn{3}{c|}{\textbf{Qwen3-4B-Base}} \\ 
\cmidrule(lr){2-4} \cmidrule(lr){5-7}  
& Acc$\uparrow$ & TC$\downarrow$ & TP$\uparrow$ & Acc$\uparrow$ & TC$\downarrow$ & TP$\uparrow$  \\
\midrule
OTC \citep{wang2025acting} & 51.52	& 1.00	& 25.76 & 49.11	& 1.00	& 24.55  \\ 
\methodname~w/ \sc{otc-st}  & \textbf{55.13}	& 1.18	& \textbf{35.37} & \textbf{51.45}	& 0.90 & \textbf{34.13}  \\
\bottomrule
\end{tabular}%
\label{tab:add_models_tool_use}
\end{table*}
\begin{table*}[t]
\centering
\small
\caption{Fine-grained tool use distribution (TC=0/1/2+ search) for baseline OTC and \methodname~w/ OTC-ST on HotPotQA. We also report answer accuracies for questions in each subset (subscript). TC=0 indicates that the model answers parametrically. \methodname~can successfully off-load questions to parametric answering (from TC=1 to TC=0) with minimal or no decrease in accuracy.} 
\setlength{\tabcolsep}{2.5pt}
\begin{tabular}{l|ccc|ccc|}
\toprule
\multirow{2}{*}{\textbf{Method}} & \multicolumn{3}{c|}{\textbf{Qwen2.5-7B-Base}} & \multicolumn{3}{c|}{\textbf{Qwen3-4B-Base}} \\ 
\cmidrule(lr){2-4} \cmidrule(lr){5-7}  
& 0 & 1 & 2+ & 0 & 1 & 2+ \\
\midrule
OTC  & $0.0_{0.0}$ & \cellcolor{blue!30}$100.0_{51.5}$ & $0.0_{0.0}$  & $0.0_{0.0}$ & \cellcolor{blue!30}$100.0_{49.1}$ & $0.0_{0.0}$  \\
\methodname~~w/ \sc{otc-st} & \cellcolor{blue!13}$34.6_{60.7}$ & \cellcolor{blue!11}$31.3_{61.3}$ & \cellcolor{blue!13}$34.1_{43.8}$  & \cellcolor{blue!15}$39.2_{53.5}$ & \cellcolor{blue!11}$31.8_{56.7}$ & \cellcolor{blue!10}$29.0_{42.9}$   \\
\bottomrule
\end{tabular}%
\vspace{-1mm}
\label{tab:added_models_search_distribution}
\end{table*} 
\begin{table*}[t]
\centering
\small
\caption{Answer accuracy and abstention classification results for specialized abstention approaches and \methodname on HotPotQA. For answer accuracy, we report overall Acc over the entire test set and Prec, i.e. accuracy over the non-abstained answers for each method. 
For classification, we report Abs(0), i.e. \% abstention for unanswerable questions (higher better), and the Delta (higher better) between the \% of abstention between unanswerable and answerable questions.}
\setlength{\tabcolsep}{4.5pt}
\begin{tabular}{l|cccc|cccc|}
\toprule
\multirow{2}{*}{\textbf{Method}} & \multicolumn{4}{c|}{\textbf{Qwen2.5-7B-Base}} & \multicolumn{4}{c|}{\textbf{Qwen3-4B-Base}} \\ 
\cmidrule(lr){2-5} \cmidrule(lr){6-9}  
& Acc & Prec & $\text{Abs}(0)\uparrow$ & Delta$\uparrow$ & Acc & Prec & $\text{Abs}(0)\uparrow$ & Delta$\uparrow$ \\
\midrule
OTC  & 0.00	& --	&100.00 &	0.00 & 0.00 &	-- &	100.00 &	0.00 \\
\methodname~w/ \sc{otc-st}  & \cellcolor{gray!20}20.98	& \cellcolor{gray!20}60.67 &	\cellcolor{gray!20}87.14	& \cellcolor{gray!20}64.51 & \cellcolor{gray!20}20.96	& \cellcolor{gray!20}53.59	& \cellcolor{gray!20}81.63	& \cellcolor{gray!20}67.07  \\
\midrule
5-shot Prompting  & 23.06	& 34.17	& 39.45	& 24.18  & 17.58 &	34.37	& 59.76	& 43.72 \\
AFH (Absolute)  & 25.52	& 36.73	& 40.3	& 27.93 & 13.44	& 47.97	& 82.81	& 41.5 \\
AFH (Multisample)  & 17.68	& 51.36	& 82.00	& 53.22 & 7.25	& 66.54	& 96.54	& 36.51 \\
DPO  & \cellcolor{gray!20}24.35 & \cellcolor{gray!20}48.25 & \cellcolor{gray!20}72.8	& \cellcolor{gray!20}60.4 & \cellcolor{gray!20}16.72	& \cellcolor{gray!20}60.84	& \cellcolor{gray!20}92.38	& \cellcolor{gray!20}75.87 \\
\bottomrule
\end{tabular}%
\label{tab:added_models_abs} \vspace{-4mm}
\end{table*}

\section{Oracle Helper Details}\label{sec:oracle_helper_appendix}
\paragraph{Implementation details} Our oracle helper experiments in Section \ref{sec:oracle_helper_main} predominantly use the same hyperparameters but differ primarily in prompts and the answer tags used in inference. During GRPO training and during warm-start synthetic data generation when $l=1$, we use the prompt described in Figure \ref{fig:oracle_helper_prompt}. Here, the $<$search$>$ tag used in normal training becomes a $<$help$>$ tag and the $<$document$>$ is replaced by $<$helper\_answer$>$. Finally, given that the message between the $<$help$>$ and $<$/help$>$ tags does not matter, we hardcode the specified ``I need help'' message during warm-start data generation when generating the help action.

\begin{figure*}
     \centering
     \vspace{4mm}
     \includegraphics[scale=0.225,trim=0mm 0mm 0mm 15mm]{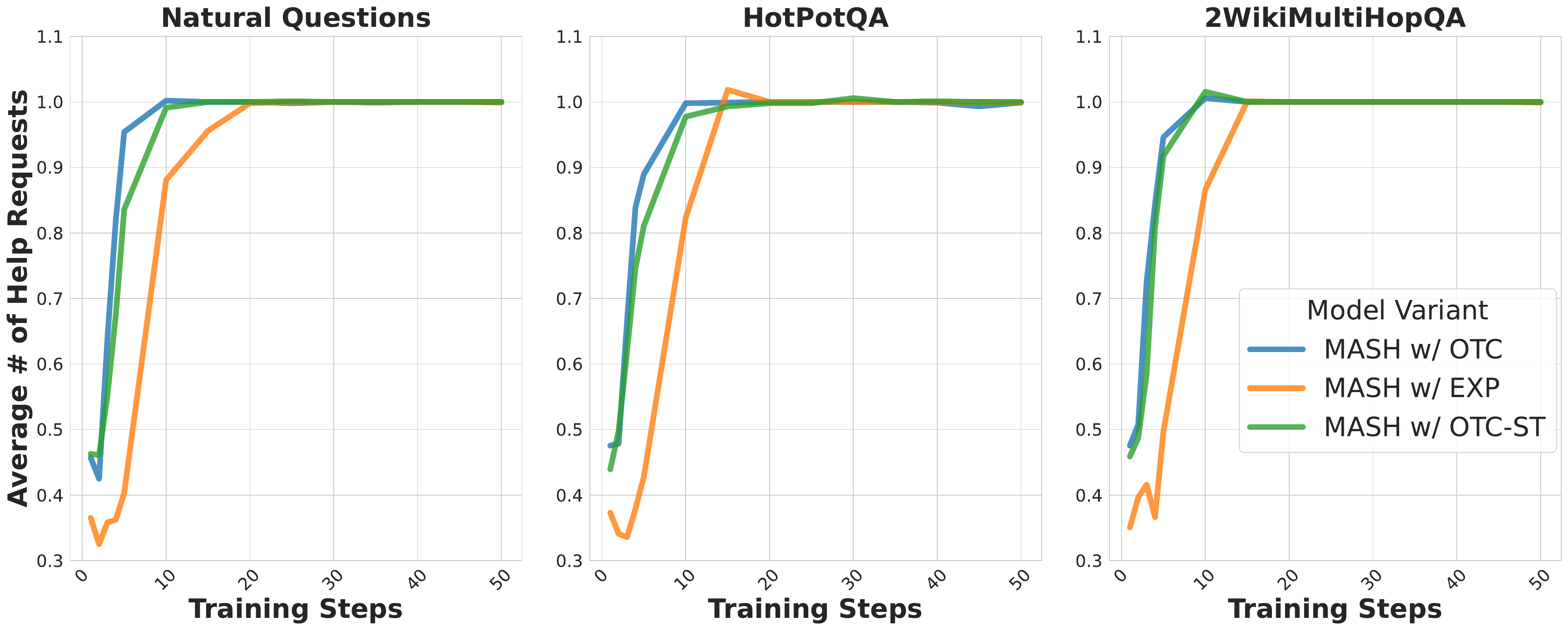}\vspace{-2mm}
     \caption{Average number of help requests for all \methodname~variants when trained with an oracle helper that directly gives away the answer. All variants converge to 1 request within 20 steps.}
     \label{fig:warm_start_hotpotqa_oracle} 
\end{figure*}

\paragraph{Visualization of help-seeking dynamics} We find that when trained with the oracle helper, all of our models, regardless of dataset, warm-start procedure or penalty severity, converge to always seeking help. Figure \ref{fig:warm_start_hotpotqa_oracle} illustrates this for MASH variants on all datasets.

\paragraph{Comparison to the ternary abstention baseline} We note that our setting with the oracle helper is equivalent to explicitly training for abstention using RL with ternary rewards. Firstly, as the oracle helper directly returns the correct answer, an efficient policy should perform at most one help call when it performs any. In practice, this results in three possible rewards with decreasing magnitude, with the maximum value given to correct answers, the intermediate value given to help seeking (abstention), and the minimum value given to incorrect answers. This is the same ternary structure explored in the Ternary baseline of Section \ref{sec:experiments}. There we observed a collapse to always abstaining. Here, we observe a collapse to always seeking help, which, under our \methodname framework is equivalent to abstention. 

\section{Additional Results}
\subsection{Additional Abstention Metrics}
While our analyses on the main paper focused on Abs(0) and Abs(1) as abstention metrics, our trained models show interpretable trends with intermediate values of Abs(0) and Abs(1). We show values for all Abs(i) values for $i \in \{0,0.1, \ldots, 0.9, 1\}$ on Tables \ref{tab:fine_grained_abs_nq} and \ref{tab:fine_grained_abs_hotpotqa}. Models' tendency to abstain decreases as the base model's average accuracy on a given question increases. We do not include results for 2WikiMultiHopQA as a majority of Abs(i) buckets do not have a high enough support.

\begin{table*}[t]
\centering
\small
\caption{Abs(i) values for each $i \in \{0, 0.1, \ldots, 0.9, 1\}$ for specialized abstention approaches and \methodname~on Natural Questions. We observe that models' tendency to abstain decreases as the average accuracy for a question increases, with a consistent drop in Abs(i) values from Abs(0) to Abs(1).}
\setlength{\tabcolsep}{2.5pt}
\begin{tabular}{l|ccccccccccc|}
\toprule
\multirow{2}{*}{\textbf{Method}} & \multicolumn{11}{c|}{\textbf{Abs(i) for Natural Questions}} \\
\cmidrule(lr){2-12}
& 0 & 0.1 & 0.2 & 0.3 & 0.4 & 0.5 & 0.6 & 0.7 & 0.8 & 0.9 & 1 \\
\midrule
OTC  & 100.00 & 100.00 & 100.00 & 100.00 & 100.00 & 100.00 & 100.00 & 100.00 & 100.00 & 100.00 & 100.00 \\
\methodname~w/ \sc{otc}  & 99.86 & 99.71 & 100.00 & 99.86 & 100.00 & 99.77 & 99.81 & 99.66 & 99.44 & 99.53 & 99.75 \\
\methodname~w/ \sc{otc-st}  & 85.45 & 75.14 & 67.27 & 52.85 & 53.62 & 51.13 & 46.40 & 33.50 & 27.78 & 27.37 & 19.29 \\
\methodname~w/ \sc{exp}  &  85.65 & 75.14 & 66.74 & 53.55 & 56.03 & 50.90 & 52.26 & 36.05 & 32.78 & 29.25 & 22.91 \\
\midrule
5-shot Prompting  & 60.16 & 49.86 & 47.31 & 38.25 & 38.45 & 31.08 & 34.09 & 25.34 & 25.37 & 21.84 & 15.60 \\
AFH (Absolute)  & 67.71 & 56.30 & 48.81 & 40.98 & 35.52 & 35.36 & 35.42 & 27.05 & 24.81 & 20.73 & 19.66  \\
AFH (Multisample)  & 87.90 & 83.41 & 78.23 & 69.95 & 70.69 & 68.92 & 60.98 & 55.65 & 52.22 & 46.68 & 35.81 \\
DPO  & 84.48 & 73.48 & 60.67 & 51.23 & 51.03 & 41.67 & 40.72 & 30.31 & 22.41 & 17.25 & 12.90 \\
\bottomrule
\end{tabular}%
\vspace{-1mm}
\label{tab:fine_grained_abs_nq}
\end{table*} 
\begin{table*}[t]
\centering
\small
\caption{Abs(i) values for each $i \in \{0, 0.1, \ldots, 0.9, 1\}$ for specialized abstention approaches and \methodname~ on HotPotQA. We observe that models' tendency to abstain decreases as the average accuracy for a question increases, with a consistent drop in Abs(i) values from Abs(0) to Abs(1).}
\setlength{\tabcolsep}{2.5pt}
\begin{tabular}{l|ccccccccccc|}
\toprule
\multirow{2}{*}{\textbf{Method}} & \multicolumn{11}{c|}{\textbf{Abs(i) for HotPotQA}} \\
\cmidrule(lr){2-12}
& 0 & 0.1 & 0.2 & 0.3 & 0.4 & 0.5 & 0.6 & 0.7 & 0.8 & 0.9 & 1 \\
\midrule
OTC  & 95.34 & 89.51 & 79.09 & 69.67 & 60.70 & 56.08 & 53.53 & 55.06 & 52.18 & 52.80 & 53.96 \\
\methodname~w/ \sc{otc}  & 94.77 & 87.43 & 75.00 & 64.35 & 52.63 & 47.95 & 41.60 & 45.62 & 44.09 & 39.20 & 42.46 \\
\methodname~w/ \sc{otc-st}  & 91.22 & 82.26 & 67.44 & 56.96 & 46.67 & 40.67 & 38.26 & 40.66 & 36.55 & 28.30 & 30.92 \\
\methodname~w/ \sc{exp}  &  94.47 & 87.25 & 74.71 & 63.28 & 52.98 & 48.03 & 41.79 & 46.80 & 43.27 & 40.24 & 41.82 \\
\midrule
5-shot Prompting  & 60.54 & 56.47 & 49.89 & 49.08 & 48.51 & 45.21 & 45.42 & 46.69 & 44.36 & 41.20 & 33.65 \\
AFH (Absolute)  &  50.42 & 42.85 & 37.50 & 32.46 & 29.12 & 27.05 & 24.43 & 22.18 & 21.18 & 19.50 & 15.03  \\
AFH (Multisample)  & 89.20 & 83.15 & 75.80 & 72.09 & 67.54 & 63.18 & 61.55 & 58.95 & 58.09 & 47.80 & 31.64 \\
DPO  & 85.91 & 70.40 & 56.28 & 47.80 & 38.07 & 33.39 & 30.06 & 26.26 & 22.64 & 20.70 & 12.44  \\
\bottomrule
\end{tabular}%
 \vspace{-1mm}
\label{tab:fine_grained_abs_hotpotqa}
\end{table*} 

\subsection{Instruct Model Prompting}
As we observed Qwen2.5-3B-Instruct to demonstrate abstention behavior off-the-shelf, we compare the performance of the OTC baseline and \methodname~w/ OTC-Strict (our best performing variant) to that of the zero- and few-shot prompted Qwen2.5-3B-Instruct model under the abstention setting. For zero-shot prompting, we use the same prompts used in inference for abstention. For few-shot prompting, we re-use the same exemplars used for few-shot prompting with the base models. 

Results can be found on \ref{tab:instruct_abstention}. We find that \methodname~w/ OTC-Strict significantly outperforms the zero- and few-shot prompted instruct model. On Natural Questions, \methodname~w/ OTC-Strict achieves a higher accuracy than the instruct model while maintaining a similar precision value. For HotPotQA and 2WikiMultiHopQA, on the other hand, the instruct model abstains for a majority of questions, resulting in a very low answer accuracy. \methodname~w/ OTC-Strict achieves substantially higher accuracies while maintaining precision in these instances.

\begin{table*}[t]
\centering
\small
\caption{Answer accuracy (left) and abstention classification (right) results for OTC, \methodname~w/ OTC-Strict and zero- and five-shot prompting for Qwen2.5-3B-Instruct. }
\setlength{\tabcolsep}{4.5pt}
\begin{tabular}{l|cc|cc|cc||cc|cc}
\toprule
\multirow{2}{*}{\textbf{Method}} &  \multicolumn{6}{c||}{\textbf{Answer Accuracy}} & \multicolumn{4}{c}{\textbf{Abstention Classification}} \\
\cmidrule(lr){2-7} \cmidrule(lr){8-11}
& \multicolumn{2}{c|}{\textbf{NaturalQA}} & \multicolumn{2}{c|}{\textbf{HotPotQA}} & \multicolumn{2}{c||}{\textbf{2Wiki}} & \multicolumn{2}{c|}{\textbf{NaturalQA}} & \multicolumn{2}{c}{\textbf{HotPotQA}} \\
\cmidrule(lr){2-3} \cmidrule(lr){4-5} \cmidrule(lr){6-7} \cmidrule(lr){8-9} \cmidrule(lr){10-11}
& Acc & Prec  & Acc & Prec  & Acc & Prec  & $\text{Abs}(0)\uparrow$ & Delta$\uparrow$ & $\text{Abs}(0)\uparrow$ & Delta$\uparrow$   \\
\midrule
OTC  & 0.0 & 0.0 & 12.6 & 64.5 & 0.75 & 24.1  & 100.0 & $0.0$ & 95.3 & 41.4 \\
\methodname~w/ \sc{otc-st}  & 20.9 & 57.4 & 17.3 & 59.9 & 4.6 & 32.5  & 85.5 & 66.2  & 91.2 & 60.3   \\
\midrule
0-shot Abstention  & 15.7 & 59.3 & 2.9 & 68.0 & 0.2 & 34.9 & 89.3 & 55.8 & 98.8 & 21.2 \\
5-shot Abstention  & 15.7 & 58.9 & 3.7 & 66.8 & 0.3 & 27.8 & 89.5 & 55.8 & 98.5 & 23.4 \\
\bottomrule
\end{tabular}%
\label{tab:instruct_abstention} 
\end{table*}

\subsection{Impact on General Task Performance}
In order to assess how our training affects the models' general capabilities, we compare Qwen2.5-3B-Base's performance against \methodname~w/ OTC-Strict on separate, general-capability tasks. We use the HotPotQA-trained variant for \methodname~w/ OTC-Strict. We compare these models on the verifiable instruction-following task of IFEval~\citep{zhou2023instruction} and the MATH-Hard~\citep{hendrycksmath2021} dataset, which features the subset of questions of the MATH dataset with level 5 difficulty. A 4-shot prompt is used in the MATH-Hard setting. We present results in Table \ref{tab:general_ability}. We do not observe any degradation in performance following our RL training. In fact, we find that our RL-trained model improves out-of-distribution, with MASH w/ OTC-Strict achieving a $1.60\%$ improvement on IFEval and an $7.68\%$ improvement on MATH-Hard.

These evaluations are done with the commonly used~\citep{gu2024mamba, touvron2023llama, muennighoff2025s1simpletesttimescaling} LM Evaluation Harness of EleutherAI~\citep{eval-harness}. We follow the standard task setup for both tasks under the LM Evaluation Harness. We use $250$ samples from each subset that is available in these datasets.

\begin{table*}[t]
\centering
\small
\caption{Performance of Qwen2.5-3B-Base and the HotPotQA-trained \methodname~w/ OTC-Strict models on IFEval and MATH-Hard. We observe training models to selectively seek help does not degrade general capabilities under our setting.}
\setlength{\tabcolsep}{2.5pt}
\begin{tabular}{l|cc|}
\toprule
\textbf{Method} & IFEval & MATH-Hard \\
\midrule
Qwen2.5-3B-Base  & 23.60 & 15.12 \\
\methodname~w/ \sc{otc-st}  & \textbf{25.20} & \textbf{22.80} \\
\bottomrule
\end{tabular}%
\vspace{-1mm}
\label{tab:general_ability}
\end{table*} 

\begin{table*}[t]
\centering
\small
\caption{Out-of-distribution accuracy (with and without search) and abstention classification results for NaturalQA models. DPO achieves superior Abs(0) and Delta, but is outperformed by \methodname~on TriviaQA. OTC consistently learns to search on NaturalQA, which generalizes out-of-distribution. However, tool-use enables both OTC and \methodname~to achieve higher accuracies.}
\setlength{\tabcolsep}{5pt}
\begin{tabular}{l|cccc|cccc}
\toprule
\multirow{2}{*}{\textbf{Method}} & \multicolumn{4}{c|}{\textbf{HotPotQA}} & \multicolumn{4}{c}{\textbf{TriviaQA}}  \\ 
\cmidrule(lr){2-5} \cmidrule(lr){6-9} 
& Acc$\uparrow$ & Acc w/ tool$\uparrow$ & $\text{Abs}(0)\uparrow$ & Delta$\uparrow$ & Acc$\uparrow$ & Acc w/ tool$\uparrow$ & $\text{Abs}(0)\uparrow$ & Delta$\uparrow$  \\
\midrule
\textsc{otc} & 0.00 & 43.05 & 99.99 & -0.01 & 0.00 & 72.51 & 99.99 & 0.01 \\ 
\methodname~w/ \sc{otc-st}  & 7.62 & 39.15 & 93.39 & 40.66 & 37.09 & 65.58 & 74.44 & 60.69
 \\
DPO   & 9.1 & - & 95.66 & 48.39 & 34.24 & - & 84.57 & 71.45  \\
\bottomrule
\end{tabular}%
\label{tab:ood_nq}
\end{table*}
\begin{table*}[t]
\centering
\small
\caption{Out-of-distribution accuracy (with and without search) and abstention classification results for 2Wiki models on HotPotQA. DPO achieves superior Abs(0) and Delta, but is outperformed by \methodname~on Accuracy. For 2Wiki, we find OTC to be more competitive with DPO than \methodname~on abstention metrics. Nonetheless, tool-use enables both OTC and \methodname~to achieve higher accuracies.}
\setlength{\tabcolsep}{5pt}
\begin{tabular}{l|cccc}
\toprule
\multirow{2}{*}{\textbf{Method}} &  \multicolumn{4}{c|}{\textbf{HotPotQA}}   \\ 
\cmidrule(lr){2-5} 
& Acc$\uparrow$ & Acc w/ tool$\uparrow$ & $\text{Abs}(0)\uparrow$ & Delta$\uparrow$   \\
\midrule
\textsc{otc} & 4.00 & 39.86 & 89.56 & 14.05  \\ 
\methodname~w/ \sc{otc-st}  & 7.07 & 39.19 & 73.36 & 17.27 
 \\
DPO   & 4.07 & - & 95.43 & 22.73   \\
\bottomrule
\end{tabular}%
\label{tab:ood_2wiki_twohop}
\end{table*}
\begin{table*}[t]
\centering
\small
\caption{Out-of-distribution accuracy (with and without search) and abstention classification results for 2Wiki models on single-hop datasets. DPO achieves superior Abs(0), but is outperformed by OTC in terms of Delta and both OTC and \methodname~in terms of Accuracy. However, we find that \methodname~struggles at abstention in this setting. Nonetheless, tool-use enables both OTC and \methodname~to achieve higher accuracies.}
\setlength{\tabcolsep}{5pt}
\begin{tabular}{l|cccc|cccc}
\toprule
\multirow{2}{*}{\textbf{Method}} & \multicolumn{4}{c|}{\textbf{Natural Questions}} & \multicolumn{4}{c}{\textbf{TriviaQA}}  \\ 
\cmidrule(lr){2-5} \cmidrule(lr){6-9} 
& Acc$\uparrow$ & Acc w/ tool$\uparrow$ & $\text{Abs}(0)\uparrow$ & Delta$\uparrow$ & Acc$\uparrow$ & Acc w/ tool$\uparrow$ & $\text{Abs}(0)\uparrow$ & Delta$\uparrow$  \\
\midrule
\textsc{otc} & 13.26 & 39.88 & 72.85 & 29.55 & 24.39 & 55.37 & 71.17 & 33.20 \\ 
\methodname~w/ \sc{otc-st}  & 11.97 & 33.34 & 40.28 & 0.05 & 23.18 & 47.42 & 49.96 & 19.44 
 \\
DPO   & 7.94 & - & 93.66 & 28.55 & 14.71 & - & 90.05 & 29.3  \\
\bottomrule
\end{tabular}%
\label{tab:ood_2wiki_onehop}
\end{table*}

\section{Out-of-Distribution Results}\label{sec:ood_appendix}
We present out-of-distribution results for models trained on NaturalQA on Table \ref{tab:ood_nq} and for models trained on 2Wiki on Tables \ref{tab:ood_2wiki_twohop} and \ref{tab:ood_2wiki_onehop}. We find that models' generalization behavior is highly dependent on the dataset they are trained on. For NaturalQA models, DPO achieves superior Abs(0) and Delta, but is outperformed by \methodname~on TriviaQA. For 2Wiki, on the other hand, where questions are exclusively multi-hop, we find that \methodname~generalizes reasonably for HotPotQA but struggles on single-hop questions. OTC, on the other hand, performs better in this setting. We note that 2Wiki is highly synthetic and that \methodname~with OTC-Strict answers parametrically $11.2\%$ more than the OTC baseline on this dataset. We suspect that \methodname~with OTC-Strict learned dataset-specific shortcuts that hamper its generalization in this process. Nonetheless, with search enabled, all of our help-seeking models outperform DPO, which is ultimately limited to abstention.

\section{Compute Requirements and Cost}
We perform all experiments on NVIDIA H100 machines. Each individual \methodname~training experiment takes approximately $100$ H100 hours for training and evaluation. In total, we perform $18$ full reinforcement learning experiments, leading to approximately $1800$ H100 hours. The various abstention experiments are cheaper due to the fact that they do not involve any retrieval, with the Alignment for Honesty Multisample training longest at approximately $4-5$ hours. Overall, we estimate all training and evaluation experiments taking approximately $1900$ H100 hours total. DeepSeek-V3.1 API calls, on the other hand, cost approximately $\$400-500$ total.

\end{document}